\pdfoutput=1
\documentclass{article}

\usepackage[numbers]{natbib}
\usepackage[utf8]{inputenc}
\usepackage[T1]{fontenc}
\usepackage{url}
\usepackage{microtype}
\usepackage{graphicx}
\usepackage{subfigure}
\usepackage{booktabs} % for professional tables
\usepackage{amsfonts}
\usepackage{nicefrac}
\usepackage{acronym}
\usepackage{microtype}
\usepackage{amsmath}
\usepackage{amsthm}
\usepackage{float}
\usepackage{adjustbox}
\usepackage{caption} 
\usepackage{thmtools, thm-restate}
\usepackage{wrapfig}
\usepackage{svg}
\usepackage{paralist}
\usepackage{mathtools, nccmath}
\usepackage{tikz}
\usepackage{xfrac}
\usetikzlibrary{arrows.meta, chains, fit, positioning}
\usepackage{algorithm}
\usepackage{listings}
\usepackage{dsfont}

\usepackage[algo2e, ruled,vlined, boxed, linesnumbered]{algorithm2e}
\usepackage{colortbl}
\SetArgSty{textnormal}

 \usepackage{enumitem}

\DeclareCaptionLabelFormat{andtable}{#1~#2  \&  \tablename~\thetable}

\makeatletter
\newcommand*{\indep}{%
  \mathbin{%
    \mathpalette{\@indep}{}%
  }%
}
\newcommand*{\nindep}{%
  \mathbin{%                   % The final symbol is a binary math operator
    \mathpalette{\@indep}{/}%
  }%
}
\newcommand*{\@indep}[2]{%
  \sbox0{$#1\perp\m@th$}%        box 0 contains \perp symbol
  \sbox2{$#1=$}%                 box 2 for the height of =
  \sbox4{$#1\vcenter{}$}%        box 4 for the height of the math axis
  \rlap{\copy0}%                 first \perp
  \dimen@=\dimexpr\ht2-\ht4-.2pt\relax
  \kern\dimen@
  \ifx\\#2\\%
  \else
    \hbox to \wd2{\hss$#1#2\m@th$\hss}%
    \kern-\wd2 %
  \fi
  \kern\dimen@
  \copy0 %                       second \perp
}
\makeatother

\usepackage[accepted, nohyperref]{icml2022}

\usepackage{amsmath}
\usepackage{amssymb}
\usepackage{mathtools}
\usepackage{amsthm}

\usepackage[capitalize,noabbrev]{cleveref}

\theoremstyle{plain}
\newtheorem{theorem}{Theorem}[section]

\theoremstyle{definition}
\newtheorem{definition}[theorem]{Definition}

\theoremstyle{remark}
\newtheorem{remark}[theorem]{Remark}
\newtheorem{conjecture}[theorem]{Conjecture}

\usepackage[textsize=tiny]{todonotes}

\pdfoutput=1
\usepackage{amsfonts}
\usepackage{amsmath}
\usepackage{bm}

\def\eqref#1{equation~\ref{#1}}
\def\1{\bm{1}}

\DeclareMathAlphabet{\mathsfit}{\encodingdefault}{\sfdefault}{m}{sl}
\SetMathAlphabet{\mathsfit}{bold}{\encodingdefault}{\sfdefault}{bx}{n}

\newcommand{\E}{\mathbb{E}}

\newcommand{\KL}{D_{\mathrm{KL}}}

\newacro{DA}{domain adaptation}
\newcommand{\DA}{\ac{DA}\xspace}

\newacro{DG}{domain generalization}
\newcommand{\DG}{\ac{DG}\xspace}

\newacro{MMD}{maximum mean discrepancy}
\newcommand{\MMD}{\ac{MMD}\xspace}

\newacro{OT}{optimal transport}
\newcommand{\OT}{\ac{OT}\xspace}

\newacro{JSD}{Jensen–Shannon divergence}
\newcommand{\JSD}{\ac{JSD}\xspace}

\newacro{KLd}{Kullback-Leibler}
\newcommand{\KLd}{\ac{KLd}\xspace}

\newacro{RKHS}{reproducing kernel Hilbert space}
\newcommand{\RKHS}{\ac{RKHS}\xspace}

\newacro{ERM}{empirical risk minimization}
\newcommand{\ERM}{\ac{ERM}\xspace}

\newacro{DAG}{direct acyclic graph}
\acrodefplural{DAGs}{direct acyclic graphs}
\newcommand{\DAG}{\ac{DAG}\xspace}
\newcommand{\DAGs}{\acp{DAG}\xspace}

\newacro{SCM}{structural causal model}
\acrodefplural{SCMs}{structural causal models}
\newcommand{\SCM}{\ac{SCM}\xspace}
\newcommand{\SCMs}{\acp{SCM}\xspace}

\definecolor{darkblue}{HTML}{1A254B}
\definecolor{lightblue}{HTML}{A7BED3}
\definecolor{blue}{HTML}{114083}
\definecolor{green}{HTML}{3e8c27}
\definecolor{pink}{HTML}{F2545B}
\definecolor{red}{HTML}{A4243B}

\DeclareMathOperator {\dist}            {dist}

\icmltitlerunning{Invariant Causal Mechanisms through Distribution Matching}

\begin{document}

\twocolumn[
\icmltitle{Invariant Causal Mechanisms through Distribution Matching}




\begin{icmlauthorlist}
\icmlauthor{Mathieu Chevalley}{yyy}
\icmlauthor{Charlotte Bunne}{yyy}
\icmlauthor{Andreas Krause}{yyy}
\icmlauthor{Stefan Bauer}{sch}
\end{icmlauthorlist}

\icmlaffiliation{yyy}{Department of Computer Science, ETH Zürich, Zürich, Switzerland}
\icmlaffiliation{sch}{Department of Intelligent Systems, KTH Stockholm, Sweden}

\icmlcorrespondingauthor{Mathieu Chevalley}{m.chevalley97@gmail.com}

\icmlkeywords{Machine Learning, ICML}

\vskip 0.3in
]



\printAffiliationsAndNotice{}  % leave blank if no need to mention equal contribution

\begin{abstract}
\looseness -1 Learning representations that capture the underlying data generating process is a key problem for data efficient and robust use of neural networks. One key property for robustness which the learned representation should capture and which recently received a lot of attention is described by the notion of invariance. In this work we provide a causal perspective and new algorithm for learning invariant representations. Empirically we show that this algorithm works well on a diverse set of tasks and in particular we observe state-of-the-art performance on domain generalization, where we are able to significantly boost the score of existing models.

%The goal of representation learning consists in learning a transformation of the data to a lower-dimensional space. A representation is often thought as a code that summarizes the latent factors of variation of the data. One goal can thus be to manipulate the learn representation to have some desirable properties, such as invariance. 
%Invariance consist in having a latent representation that do not vary accodring to some random variable. 
%In this work, we first give a more precise definition of invariance. We then argue that many common machine learning task can be cast as an invariant representation learning task. We propose a \SCM to model the generation process of the data, and we argue that most invariant representation learning datasets follows our proposed assumptions. We then develop some theory and prove some necessary and sufficient conditions for enforcing invariance towards unobserved latent factors, that we call style variables. Finally, we propose a new algorithm to enforce invariance based on our assumptions on the data generation process. We empirically show that this algorithm works on a diverse set of tasks, which confirms the validity of our assumptions. We also observe state-of-the-art performance on \DG, where we are able to significantly boost the score of existing models with our algorithm.
\end{abstract}

%%%%%%%%%%%%%%%%%%%%%%%%%%%%%%%%%%%%%
%
%
%
%%%%%%%%%%%%%%%%%%%%%%%%%%%%%%%%%%%%

\section{Introduction}

\looseness -1 Learning structured representations which capture the underlying data-generating causal mechanisms  is of central
importance for training robust machine learning models \citep{bengio2013representation,scholkopf2021toward}. One particular structure the learned representation should capture is invariance to changes in nuisance variables. 
%In today's state of Machine Learning research, most state-of-the-art models for common tasks are based on Deep Learning. Deep Learning models are especially powerful for structured data such as images, sounds and medical measurements. Their success has been largely attributed to their capacity to learn meaningful \emph{representations} of data \citep{bengio2013representation}. Representations are useful for multiple reasons. They provide lower dimensional features for the data on which it is easier to train a downstream model, such as a classifier or even a Reinforcement Learning agent. They can be reused across different tasks. They can also demonstrate some more refined properties, such as obfuscating part of the information contained in the data. Current Deep Learning research thus mainly focuses on finding new inductive biases to shape and steer the learned representation to theses desired properties. %One such properties is the idea that the representation should be insensitive to variations of a variable.
For example, we may want the representation to be \emph{invariant} to sensitive attributes such as the race or gender of an individual in order to avoid discrimination or biased decision making in a downstream task \citep{creager2019flexibly,locatello2019fairness, trauble2021disentangled}.

%One possible inductive bias to enforce this property is to add an adversary during training. This adversary tries to predict the sensitive attribute given the representation as input. The goal of the encoder, which output the representation, is then to fool this adversary. As it was shown in the seminal work on adversarial training \citep{Goodfellow2014}, the asymptotic way for the encoder to completely fool the adversary is to output a representation whose statistical distribution is independent to the sensitive attribute. This allows us to abstract the inductive bias of adversarial training further, by reformulating it as \emph{the distribution of the representation should be the same across different values of the sensitive attribute}. This idea of invariant representation is obviously not new \citep{muandet2013domain, pmlr-v28-zemel13, peters2016causal}. What we here want to emphasis is that the inductive bias of adversarial training is not different from distributional matching, and that they should be treated similarly. 

%With this observation that many commonly used inductive bias are intrinsically equivalent to enforcing an invariant distribution of the representation, we can now ask us in which cases and under which condition this makes sense.

%Furthermore, invariant representation via invariant distribution of the representation also appears in seemingly unrelated tasks. 

\looseness -1 While learning invariant representations is thus highly important for fairness applications, it also appears in seemingly unrelated tasks such as \DA and \DG, where one aims to be invariant across the different domains
\citep{muandet2013domain, pmlr-v28-zemel13, ganin2016domain, peters2016causal}.  For tasks such as \DA and \DG, invariance across domains or environments implies being invariant to the domain index, which thus is the ``sensitive attribute'' in this case  and typically implies a change in the distribution of the data generating process. %Nevertheless, being invariant to the domain index is somewhat arbitrary and is only based on intuition. Typically the split between different domains is manually decided and can thus be considered as an inductive bias from the researcher who designed the dataset. Intuitively, the domain index denotes a change in the distribution of the data. 
Being invariant to the domain index is thus a proxy to being invariant to latent unobserved factors that can change in distribution.

%Intuitively, style factors are generating factors that should not be used to perform predictions. What is a style factors and what is a content factor depends on the task and application. For example, in \emph{sim-to-real} tasks, i.e $D = \{simulated, real\}$, we want to learn a model that for instance does not rely on the texture or brightness of an image, as these factors may change in distribution between the simulated and real setting. Unfortunately, as we may not have access to these unobserved factors, we can only use the division of the dataset between $simulated$ and $real$ samples as a proxy/learning signal.

%\stefan{The sim to real example is good as motivation, I am just not sure if we want to keep it given that we should not include the experiments for this}

\looseness -1 Established approaches for enforcing invariance in the learned representation usually aim to learn a representation whose statistical distribution is {\em independent} of the sensitive attribute, e.g., by including an adversary during training \citep{ganin2016domain, xie2017controllable}. As an adversary is essentially a parametric distributional distance, other approaches minimize different distribution distances, such as \MMD \citep{louizos2017variational, li2018domain}, or \OT based distances \citep{shen2018wasserstein, damodaran2018deepjdot}. To enforce independence, these methods add a regularizer to the loss that consists of the pairwise distributional distance between all possible combinations of the sensitive attribute, i.e., $\dist(p(z|d), p(z|d')) \forall d, d' \in D$. As such, the complexity of the loss grows quadratically in the size of the support of the sensitive attribute, which can limit the applicability of these models when the support of $D$ is large \citep{wilds2021}.%these models have been observed to suffer when the support of $D$ is too large. 

%Despite the importance of learning invariant representations and their potential societal impact in the medical domain or fair decision making, most established approaches are typically based on heuristics and specialized for different tasks at hand. 

Despite the importance of learning invariant representations and their potential societal impact in the medical domain or fair decision making, a commonly accepted definition is still missing and most established approaches are specialized for different tasks at hand. We take first steps towards a unifying framework by viewing \emph{invariant representation learning} as a property of a causal process \citep{pearl2009causality, peters2017elements} and our key contributions can be summarized as follows: 

%In this work we view \emph{invariance} as a property of a causal process \citep{pearl2009causality, peters2017elements} which offers a common framework of invariant representation learning through invariant latent variables distributions.
%We introduce a unifying causal framework for invariant representation learning, which allows us to derive a new algorithm to enforce invariance through distribution matching. We define the notion of style variable and present some necessary and sufficient conditions under which being invariant to the domain index actually leads to invariance to the style variables. We argue that our proposal naturally captures most of the existing invariant representation learning tasks and datasets. Finally, we conduct a large number of experiments across different tasks and datasets, demonstrating the versatility of our framework. We obtain competitive results on the task of learning fair representations and we are able to significantly boost the performance of existing model using our proposed algorithm for the task of \DG. \textcolor{blue}{One advantage of our algorithm is that only one distributional distance between two batches needs to be computed at each step, irrelevant of the size of the support of $D$.}

\begin{itemize}[noitemsep]
    \item We introduce a unifying framework for invariant representation learning, which allows us to derive a new simple and versatile regularizer to enforce invariance through distribution matching. One advantage of our algorithm is that only one distributional distance between two batches needs to be computed at each step, irrelevant of the size of the support of $D$. 
    \item By enforcing a softer form of invariance, our proposed method offers a new tool with a better trade-off between predictability and stability.
    \item Finally, we conduct a large number of experiments across different tasks and datasets, demonstrating the versatility of our framework. We obtain competitive results on the task of learning fair representations and we are able to significantly boost the performance of existing models using our proposed algorithm for the task of \DG. 
    %argue that many different tasks are inherently related to invariant representation learning, and we propose a data generation framework that aims to unify them. 
    %\item Based on the introduced framework, we propose a new algorithm and theoretic insights on necessary and sufficient conditions on enforcing invariance when learning representations from data. 
    %\item We also define the notion of style variables in the context of our framework.
   % \item We develop some theory to gain insights on whether and when invariance to the sensitive attribute may lead to invariance to the style variables.
   
   % \item We review the literature on different invariant representation learning tasks and connect previous works under our framework.
    %\item We conduct a large number of experiments across different tasks and datasets, demonstrating the versatility of our framework. We obtain competitive results on the task of learning fair representations and we are able to significantly boost the performance of existing model using our proposed algorithm for the task of \DG. 
\end{itemize}

\section{Invariant Representation Learning Across Tasks}
\label{chp:review}

In this section, we highlight how the learning of an invariant representation is a goal that is (implicitly) pursued in a large spectrum of machine learning tasks.  %We will see that many methods and models are reinvented in different lines of research, as many machine learning tasks consist in being invariant to certain variables, these variables being different across tasks (e.g, a sensitive attribute in fair representation learning, the dataset index in domain adaptation and domain generalization, a transformation in contrastive learning). The difference between these tasks is then only a matter of the datasets used, and each task presents different engineering and optimization challenges.

\paragraph{Domain Generalization}
The task of \DG seeks to learn a model that generalizes to  an unseen domain, given some domains at training time. As such, it is a very harder task as the test domain could exhibit arbitrary shifts in distribution, and the learned model is supposed to handle any \emph{reasonable} shifts in distribution. Without any assumptions, there is little hope to obtain models that actually generalize. Nevertheless, many inductive biases and models have been proposed, which have stronger assumptions than classical \ERM \citep{vapnik1998statistical}. % Contrary to Domain Adaptation, most benchmark datasets consists of multiple training domains, which should guide the learning of a representation that keeps what is constant (the content) and discard what is varying (the style). One such dataset is the DomainNet \citep{peng2019moment} dataset, where objects are presented under different styles, such as real photographs, paintings, quick draws etc. Such datasets implicitly assume the existence of a function that can transform an object across domains.

Given its similarity to \DA, similar models have been proposed, and most models work for both tasks. Nevertheless, until recently \citep{albuquerque2019generalizing, deng2020representation}% \stefan{If you say recently then there should be a citation for this? You mean then the paper from Deng you cite late?}
, theoretical justification, e.g., for minimizing the distance between pairs of latent variables coming from different domains, was missing, as results from domain adaptation assume that the test domain is observed. Without some assumptions, there exists no theoretical reasons to infer that a constant distribution of the latent variables across the training domains leads to better generalization on the test domains. Indeed, many benchmarks \citep{gulrajani2020search, wilds2021} show that it is difficult to create algorithms that consistently beat \ERM across different tasks. Invariant representations for \DG was first proposed by \citet{muandet2013domain}. This idea was then extended to use other distributional distances, such as MMD \citep{li2018domain}, Adversarial \citep{li2018deep, deng2020representation, albuquerque2019generalizing}, and Optimal Transport \citep{zhou2020domain}. %(see \cref{tab:review_invariance})
 On the theoretical side, both \citet{albuquerque2019generalizing} and \citet{deng2020representation} attempt to give theoretical grounding to the use of an adversarial loss by deriving bounds similar to what exists in \DA. 
%\vspace{-0.25em}
\paragraph{Domain Generalization and Causal Inference}
\label{sec:dg_ci}

Many links between causal inference and domain generalization have been made, arguing that domain generalization is inherently a causal discovery task. In particular, causal inference can be seen as a form of distributional robustness \citep{causalitymein}. In regression, one way of ensuring interventional robustness is by identifying the causal parents of $Y$, whose relation to $Y$ is stable. This can be achieved by finding a feature representation such that the optimal classifiers are approximately the same across domains \citep{peters2016causal, rojas2018invariant}. Unfortunately, most of these models do not really apply to classification of structured data such as images, where the classification is predominantly anti-causal and where the wanted invariance is not toward the pixels themselves but towards the unobserved generating factors. In a similar setting to ours, \citet{heinze2021conditional} tackle the task of image classification and propose a new model. A significant difference to our work is that they rely on the observation of individual instances across different views, i.e., the images are clustered by an ID. 

\paragraph{Fair Representation Learning}

\looseness -1 Fair representation learning can also be viewed as an invariant representation learning task. This task seeks to learn a representation that maximizes usefulness towards predicting a target variable, while minimizing information leakage of a sensitive attribute (e.g., gender, age, race). The seminal work of \citet{pmlr-v28-zemel13} aims at learning a multinomial random variable $Z$, with associated vectors $v_k$, such that the representation $Z$ is fair. More recent work directly learns a continuous variable $Z$ that has minimal information about the sensitive attribute, either through minimizing the \MMD distance \citep{louizos2017variational}, through adversarial training \citep{edwards2015censoring, xie2017controllable, roy2019mitigating}, or through a Wasserstein distance \citep{jiang2020wasserstein}. %Most fairness datasets can be argued to follow our assumption \stefan{We have not explained the assumption yet?}, as the sensitive attribute can often be considered to have no causal parents (e.g the age or sex of an individual is not the effect of another variable).

\iffalse

\begin{table*}[t]
\centering
\caption{Review of invariance across different tasks. Note that the general loss is defined as $\frac{1}{n} \sum_{i=0}^n \mathcal{L} (x_i, y_i) + \lambda \cdot \left ( \dist(z_1^n, z_{n + 1}^N) \right ) $.}
\label{tab:review_invariance}
    \vskip 0.15in
\begin{center}
\begin{tiny}
\begin{sc}
    \begin{tabular}{llll}
    \toprule
         \textbf{Task} &  \textbf{Adversarial} & \textbf{MMD} & \textbf{Wasserstein} \\
         \midrule
          Distance equation & $\frac{1}{n} \sum_{i=0}^n \log \frac{1}{G_d(z_i)} + \frac{1}{n'} \sum_{i=n+1}^N \log \frac{1}{1 - G_d(z_i)}$ & $\left \| \frac{1}{n} \sum_{i=0}^n \phi(z_i) - \frac{1}{n'} \sum_{i=n+1}^N \phi(z_i) \right \|_{\mathcal{H}}$ & $\frac{1}{n} \sum_{i=0}^n G_d(z_i) - \frac{1}{n'} \sum_{i=n+1}^N G_d(z_i)$\\
         \midrule
         Domain Adaptation &  \cite{ganin2016domain} & \cite{dammd} & \cite{shen2018wasserstein}\\
         & \cite{hoffman2017cycada} & & \cite{damodaran2018deepjdot}\\
         Domain Generalization & \cite{ganin2016domain, albuquerque2019generalizing} & \cite{li2018domain} & \cite{zhou2020domain} 
         \\
         & \cite{li2018deep, Li_2018_ECCV, deng2020representation} & & \\
         Fair Representation Learning & \cite{edwards2015censoring, xie2017controllable} & \cite{louizos2017variational} & \cite{jiang2020wasserstein}\\
         & \cite{roy2019mitigating} & & \\
         \bottomrule
    \end{tabular}
\end{sc}
\end{tiny}
\end{center}
\vskip -0.1in   
    
\end{table*}

\fi
\newcommand{\X}{\mathcal{X}}
\newcommand{\Y}{\mathcal{Y}}
\newcommand{\aB}{\mathbf{a}}
\newcommand{\bB}{\mathbf{b}}
\newcommand{\one}{\mathds{1}}

\section{Background in Causality}
\label{cha:background}

\looseness -1 In this section, we review some necessary theoretical background that is required for the introduction of our framework and the motivation for our newly proposed algorithm. %We introduce the main elements of causality theory, which is the central theoretical basis used to model the problems we study and to design new methods and algorithms. 
In \cref{app:back}, we also review some distributional distances used in this work, as our main goal is to study invariant representation learning via invariant latent variable distributions.

%\paragraph{Causality}

Causality essentially is the study of cause and effects, which goes beyond the study of statistical associations from observational data. This allows to reason about the notion of \emph{interventions}, such as a treatment in medicine. The expected effect of an intervention is in general not equivalent to statistical conditioning, which calls for a more profound understating of the data generating process that goes beyond correlations between variables. We here focus on Pearl's view of causality \cite{pearl2009causality}, which mainly relies on \DAGs.

A \DAG allows to represent the relations between variables, where each variable is represented by a node in the graph. Consequently, we can interpret directed edges between nodes as the existence of a causal effect from the parent node (the cause) on the child node (the effect). 

Let $G = (V, E)$ be a \DAG, and $P$  be a distribution. We say that $(G, P)$ is a {\em causal \DAG model} if for any $W \subset V$ we have:

\begin{align*}
    p(x_V | do(X_W = x'_W)) & = \prod_{i \in V \backslash W} p(x_i | x_{pa_i}) \mathrm{I}(x_W = x'_W)
\end{align*}

\looseness -1 where $x_{pa_i}$ are the parents of node $i$ in graph $G$, $\mathrm{I}$ is the indicator function and $do(X_W = x'_W)$ denotes the intervention on the variables $X_W$. As we can see above, one of the properties of a causal \DAG model is that the distribution factorizes according to the parents in the associated graph~$G$.

\paragraph{Structural Causal Models}

A \SCM can be seen as a more expressive version of a causal \DAG model. Formally, an \SCM consists of a collection $S$ of $d$ structural assignments, one per variable: 

\begin{align*}
    X_j \xleftarrow{} f_j(X_{pa_j}, N_j)
\end{align*}

where $X_{pa_j} \in X \backslash X_j$, and $N_1$ to $N_d$ are called the \emph{noise variables} \citep[Definition 6.2 of][]{peters2017elements}. The noise variables are assumed to be jointly independent.  

For causal \DAG models, we defined an intervention by $p(X_V | do(X_W = x'_W))$ (sometimes also written as $p^{do(X_W = x'_W)} (X_V)$), where the value of some variables are set to a constant value. With \SCMs, we can give a more precise and general definition of interventions. An intervention now consists of replacing a subset of the collection $S$ of structural assignments by new functions. An intervention can thus consist of replacing a variable by a constant, a new random variable or even by changing the function and its arguments (i.e., its parents). The new distribution over the variables entailed by the intervened \SCM is denoted by $P^{do \left (X_k = \tilde{f}(X_{\widetilde{pa}_k}, \tilde{N}_k) \right )} $ \citep[for more details see Definition 6.8 of][]{peters2017elements}.

With this definition, we can now present the important notion of \emph{Total Causal Effect}. 

\begin{definition}
\label{def:total_causal_effect}
\citep[Definition 6.12 in][]{peters2017elements} We say that a variable $i$ has a total causal effect on a variable $k$ if and only if:

\begin{align*}
    X_i \nindep X_k \text{  in  } P^{do \left (X_k = \tilde{N}_k \right )}\\
\end{align*}

for some random variable $\tilde{N}_k$.

\end{definition}

A total causal effect between a variable $X_k$ and $X_i$ may only exist if there is a directed path from $i$ to $k$ in the \DAG associated to our \SCM. On the other hand, there may be no total causal effect between two variable even though there exists a directed path between them in the graph.

\section{Invariance as the Property of a Causal Process}

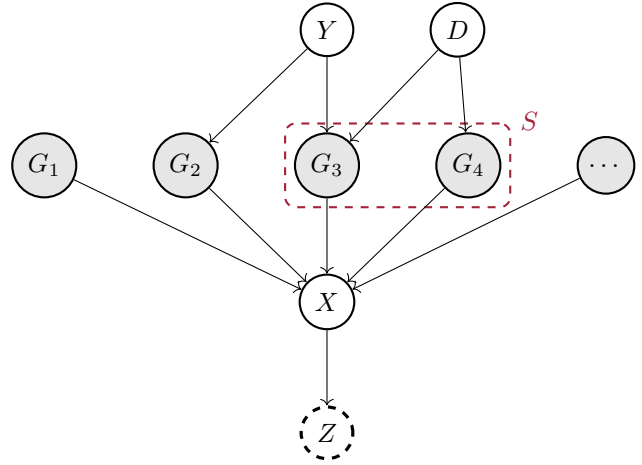
\begin{figure}[ht!]
\vskip 0.2in
\begin{center}
\centerline{\begin{tikzpicture}[
%transform canvas={scale=0.8},
%scale=0.8,
%every node/.style={scale=0.8},
roundnode/.style={circle, draw=black, thick},
roundnodegrey/.style={circle, draw=black, fill=gray!20, thick},
roundnoded/.style={circle, draw=black, dashed, very thick},
]
%Nodes
\node[roundnode]      (y)                              {$Y$};
\node[roundnode]        (d)       [right=of y] {$D$};
\node[roundnodegrey]      (g3)       [below=of y] {$G_3$};
\node[roundnodegrey]      (g4)       [right=of g3] {$G_4$};
\node[roundnodegrey]      (g2)       [left=of g3] {$G_2$};
\node[roundnodegrey]      (g1)       [left=of g2] {$G_1$};
\node[roundnodegrey]      (dot)       [right=of g4] {$\dots$};
\node[roundnode]      (x)       [below=of g3] {$X$};
\node[roundnoded]      (z)       [below=of x] {$Z$};

\node[draw, thick, color=red, dashed, rounded corners, fit=(g3) (g4)] (box) {};
\node[color=red]      (s)      [above right=10pt and 0pt of box.east] {$S$};

%Lines
\draw[->] (y) -- (g2);
\draw[->] (y) -- (g3);
\draw[->] (d) -- (g3);
\draw[->] (d) -- (g4);
\draw[->] (g1) -- (x);
\draw[->] (g2) -- (x);
\draw[->] (g3) -- (x);
\draw[->] (g4) -- (x);
\draw[->] (dot) -- (x);
\draw[->] (x) -- (z);
\end{tikzpicture}}
\caption{A \DAG exhibiting our assumptions on the data generating process. We suppose that the data $X$ is a function of unobserved generative factors $G$ %(e.g., background colors, brightness, noise, shape)
. There may exist some confounders $Y$ and $D$ that are parents of the generating factors. $Y$ is the variable that we want to predict. $D$ is the variable we want to be invariant to. Only $X$, $D$ and possibly $Y$ are observed at training time. The representation variable $Z$ is a function of the data $X$ that we \emph{create} at training time. %\stefan{Make this image smalle and only half of the page through wrapfigure for example} 
}
\label{fig:graph}
\end{center}
\vskip -0.2in
\end{figure}

In this section, we first consider the assumptions for the causal process underlying the data generating mechanism using a \SCM type graph from Causality theory \citep{pearl2009causality} and following the causal view of learning disentangled representations \citep{suter2019robustly}, as illustrated in \cref{fig:graph}.

%The goal of this graph is that it presents a general enough framework that covers most common tasks in Deep Learning. It especially concentrates on structured data such as images. The idea is that with the right assumptions, we may be able to explain and justify some commonly used inductive biases, in particular the invariant representation assumption. 
%We take inspirations from the disentangled representation literature, and the different works that follows a causal view of it \citep{suter2019robustly}. %The main purpose of our work is thus to assess whether this framework accurately captures the underlying data generation process, to use these assumptions to prove new theoretical results and gain insights on commonly used inductive biases.
%\subsection{Structural Causal Model}
%We present our data generation framework with the Structural Causal Model graph in \cref{fig:graph}. 
$G_1$ to $G_k$ represent all the factors of variation that generate the data, i.e., there exists a (one-to-one) function such that given all the factors, $X$ is fixed: $ X \xleftarrow{} g(G_1, \dots, G_k)$.

$Y$ is a target value that we may want to predict in a downstream task and is either known (supervised setting) or unobserved (unsupervised). %We thus focus on anti-causal predictions, and we argue that most common tasks, especially in computer vision, are anti-causal (e.g. classification, diagnosis).
$D$ is another confounder that we want to be invariant to. It can be a domain index, such as in \DA and \DG, or a sensitive attribute such as in fairness. We will assume for now that $D$ does not have an effect on $Y$.

Lastly, the generative factors $G_i$ are assumed to not have any causal relations between them, and any correlation between some factors may only come from a hidden confounder. This assumption is similar to the assumptions of \citet{suter2019robustly}.
%Our view of the data generating process is particularly suitable for structured data such as images. Indeed, in that setting, reasoning about the interaction between different pixels does not really make sense. On the other hand, if we assume the existence of a set of generative factors (e.g shape, color, orientation, brightness) that fully identifies an image, we can then reason about the interactions and correlations between these factors. 
Furthermore, in this work, we assume that  the label $Y$ and $D$ directly have an effect on the latent generating factors. In this setting, $Y$ and $D$ are thus independent. 
%We can think of settings where this assumptions does not hold, for example when $Y$ is an animal species and $D$ is the country where the image was taken. Here, we can see that the country has an effect on the distribution of types of animals. Nevertheless, being invariant to the country could improve robustness, as it could remove unwanted correlations such as the model relying on the background (e.g. an elephant on a beach may be missclasified if this configuration does not exist in the training set). We will not dive into this direction of having causally related $Y$ and $D$, and leave it as potential future work.

Given our data generating framework, we can now give some definitions, especially the notion of style generating factors. 

\begin{definition}
\label{def:style_var}
    We call {\em style variables} the set of variables $G$ that are children of $D$ in the \DAG. We denote this set $S$. %The rest of the variables are called content variables, $C = G \backslash S$.
\end{definition}

\begin{remark}
    $X$ and $Z$ are independent from $D$ given $S$, as they are $d$-separated from $D$ by the set $S$ in the graph. This implies that independence to $D$ is a necessary condition for $Z$ to be independent from $S$.
\end{remark}

 In this work, we propose and use the following definition of an invariant representation:
 
 % and we argue that the following proposed definition based on intervention may be strong but is also very robust.

\begin{definition}
    We say that a representation $Z$ is \emph{invariant} to a variable $D$ if and only if $D$ has no total causal effect %\footnote{A variable $i$ has a total causal effect on a variable $k$ if and only if: $X_i \nindep X_k \text{  in  } P^{do \left (X_k = \tilde{N}_k \right )}$ for some random variable $\tilde{N}_k$ (Definition 6.12 in \cite{peters2017elements}).
    %i.e. they do not effect each other as defined in \ref{def:total_causal_effect}
    %}
    on $Z$ (\cref{def:total_causal_effect}).
\end{definition}

%\stefan{Why is this definition of conditional alignment here? Do we even need it anywhere?}
%We also present the following definition of conditional alignment. Indeed, invariance to $D$ may not be sufficient to have a representation that is useful for downstream tasks.

%\begin{definition}
%    Conditional alignment: the distribution of $z$ is conditionally aligned if for all $c \in C$, $S$ has no total causal effect on the mechanism $p(z | c)$.
%\end{definition}

%Having conditional alignment ensures that models learned on the representation are transferable across changes in distribution of the style variables. 

The goal of invariant representation learning can then be described as creating a new variable $Z = f(X)$ such that $D$ has no total causal effect on $Z$. In a way, we can view it as adding a new variable in the \SCM and learning its structural equation.

\iffalse
\begin{restatable}{theorem}{thmequivalence}
\label{thm:equivalence}
Under the assumption of the graph in \cref{fig:graph}, we have that:
    
    $Z$ is independent from $D$ (equivalently, $D$ has no total causal effect on $Z$, or $p(z|d) = p(z|d')$ for all $d, d'$) $\iff p(z|do(d = N_d)) = p(z)$ for all $N_d$ (intervention on the distribution of $D$).
\end{restatable}
\fi

\begin{remark}
\label{rem:equiv}
    Invariance of $Z$ from $D$ can be enforced in two equivalent ways: either $p^{do(D = d)} (Z) = p (Z | D = d)$ to be constant for all $d$ (as D has no causal parents in our graph, intervention is equivalent to conditioning), or $p^{do(D = \tilde{N}_D)} (Z) = \int p (Z | D = d) \tilde{N}_D(d) \mathrm{d} d$ to be constant for all $\tilde{N}_D$. That is, we can enforce invariance through hard or soft interventions on $D$.
\end{remark}

\looseness -1 By being invariant to $D$, we are implicitly trying to be invariant to the style variables $S$, whose distributions are unstable. We argue that there thus exists a trade-off between predictive power of the representation and invariance, which comes from the fact that some generative factors are children of both $Y$ and $D$. Being strongly invariant to $D$ (and $S$) may then be detrimental to performance. This is likewise indicated in recent other works from the causality literature which show the benefits of models which trade-off stability and predictiveness \citep{basu2018iterative, pfister2019learning, pfister2021stabilizing, rothenhausler2021anchor} over models which are only purely focused on achieving invariance \citep{peters2016causal}.

\section{An Algorithm for Invariant Latent Variable Distributions}
\label{cha:theory}

\begin{figure}[ht!]
\vskip 0.2in
    \begin{center}
    \centerline{\includegraphics[width=0.9\columnwidth]{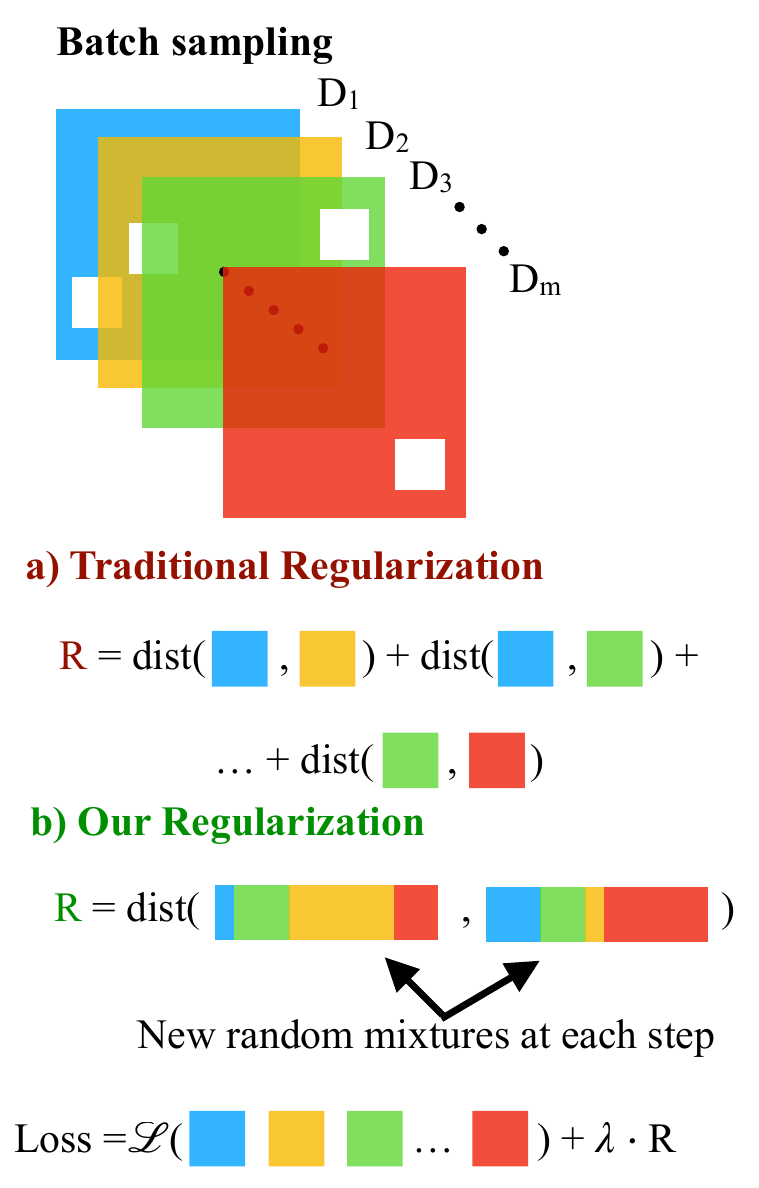}}
    \caption{Visual representation of our proposed algorithm and regularization. For $\lambda = 0$, we recover traditional ERM. To compute the regularization, any distributional distances ($\dist$ in the figure) can be used. See \cref{app:back} for a review of possible distances. At each step of optimization, batches (colored squares) are drawn for each value of $D$ (e.g, from each domain). Those batches are then encoded and a distance between latent codes is computed for the regularization R. Traditionally, the regularization \textcolor{red}{R} is computed by taking the distance between pairs of batches of latent codes coming from different domains. Instead, we propose to compute \textcolor{green}{R} by taking only one distance between mixtures of latent codes coming from different domains. The distribution of the mixtures is changed at each step. }
    \label{fig:algo}
    \end{center}
\vskip -0.2in
\end{figure}

%We now present a new algorithm for learning a representation invariant to $D$, 
Based on the underlying assumptions of \cref{fig:graph} and \cref{rem:equiv}, we present a new algorithm to learn a representation invariant to soft interventions on $D$. This algorithm could be useful for example when we have a large number of different values of $D$, where enforcing an invariant $p(z|d)$ is hard to optimize (pairwise distances between distributions). Instead, we change the distribution of $D$ across batches (simulated soft intervention) and take the distribution distance between pairs of batches. %It also follows \cref{cor:intervention}: as each time we create a batch with a different distribution of $D$, it is equivalent to drawing a batch from a created domain, which is a mixture of the initially given domains. 
We formulate the optimization goal as follows:
\begin{equation}
\label{eq:primal_1}
\begin{aligned}
    \min_{Z = f(X)} \mathcal{L}(Y, c(Z)), \enspace \\
    \text{ s.t. }p(Z) = \text{const }\forall N_d. 
\end{aligned}
\end{equation}

Now, let $Q$ be a probability measure with full support over distributions $N_d$ that have full support over $D$. We reformulate \cref{eq:primal_1} as follows:

\begin{equation}
\label{eq:primal_2}
\begin{aligned}
    & \min_{Z = f(X)} \mathcal{L}(Y, c(Z)), \enspace \\
    & \text{ s.t. } \E_{N_d, N'_d \sim Q} \left [\dist(p^{do(d = N_d)}(Z), \enspace p^{do(d = N'_d)}(Z)) \right ] = 0,
\end{aligned}
\end{equation}

where $\dist$ is a distance between distributions (see \cref{cha:background} for possible distances), and $N_d, N'_d$ are interventions on the distribution of $d$ drawn from $Q$.

For infinite data, the two constraints are equivalent: the first constraint trivially implies the second. For the other direction, the expectation being $0$ implies $\dist(p^{do(d = N_d)}(Z), \enspace p^{do(d = N'_d)}(Z)) = 0$ almost surely, as a distance function is non-negative. Lastly, there always exists a solution $f$ that satisfies the constraint, e.g, $f(x) = c \enspace \forall x, \text{with } c \in \mathbb{R}$.

We then relax this constraint by taking the dual formulation:

\begin{equation}
\label{eq:loss}
\begin{aligned}
    &\min_{Z = f(X)}  \mathcal{L}(Y, c(Z)) \enspace + \\
      &\lambda \cdot \E_{N_d, N'_d \sim Q} \left [\dist(p^{do(d = N_d)}(Z), \enspace p^{do(d = N'_d)}(Z)) \right ].
\end{aligned}
\end{equation}

%\stefan{where $N_d$ is again the interventions on $D$. Repeat that here sucht that they do not have to search for the definition.}

%First, this algorithm gives us a new method to learn invariance when our dataset follows the assumption of \cref{fig:graph}. If the dataset indeed fulfills our assumptions, we theoretically have that the algorithm will work asymptotically. % \textcolor{red}{Can you add a theorem for this? and be it in the appendix. but would actually put it here.}% Unfortunately, these assumptions are not testable. There is thus two possible outcomes: either the optimization converges and we cannot reject that the dataset follows our assumptions, or it does not converge and it probably means that the causal relationships of the variables of our dataset are different than assumed.

%Second, as stated in \cref{cor:intervention}, being invariant to different distributions of $D$ may lead to greater invariance to the style variables. This is especially useful in \DG. In \cref{cha:experiments}, we experimentally show that our algorithm is indeed a viable method to learn invariance, and that it can also be more favorable in settings such as \DG. We present a possible practical implementation of our \cref{alg:invariance} below.
This algorithm gives us a new method to learn invariance. Furthermore, as we minimize the average distance between soft interventions, we intuitively impose a softer regularization than taking a distance between hard interventions. This may lead to models that exhibit a better trade-off between invariance and predictive power of the representation.

\begin{algorithm2e}
\SetAlgoLined
Let $d$ be the number of domains\;
Let $n > 0$ be the number of samples drawn from each domain at each step\;
\Begin{
    Draw a batch $b_i$ of $n$ samples for each domain\;
    $B1, B2 \gets \emptyset, \emptyset$\; \tcp{We create two batches $B1$ and $B2$ that approximate the interventions $N_d$ and $N'_d$ of \cref{eq:loss}}
    \For{$i \gets 1$ to $d$} {
      $s \sim \mathcal{U}(0, n)$\;
      $B1, B2 \gets (B1, b_i[:s]), (B2, b_i[s:])$\; \tcp{Concatenate B1 and B2 with a slice of $b_i$}
    }
    $Z1, Z2 \gets f(B1), f(B2)$\;
    $loss \gets \dist(Z1, Z2)$\;
    \Return $loss$\;
 }
 \caption{Our algorithm for invariant representation learning.}
 \label{alg:invariance}
\end{algorithm2e}

\section{Empirical Evaluation} % or Applications
\label{cha:experiments}

\subsection{Synthetic Experiment}

We first conduct a simple synthetic experiment to verify that our algorithm effectively enforces invariance to $D$ in a setting that exactly follows our assumptions. We also simplify the setting by considering that we directly observe the generative factors.

\iffalse
\begin{figure}[h!]
    \centering
    \begin{tikzpicture}[
roundnode/.style={circle, draw=black, thick},
]
%Nodes
\node[roundnode]      (x1)   {$G_1$};
\coordinate [right =of x1] (c1);
\node[roundnode]      (x2)       [right =of c1] {$G_2$};
\coordinate [right =of x2] (c2);
\node[roundnode]      (x3)       [ right=of c2] {$G_3$};
\node[roundnode]      (y)    [above=of c1]     {$Y$};
\node[roundnode]        (d)       [above=of c2] {$D$};

%Lines
\draw[->] (y) -- (x1);
\draw[->] (y) -- (x2);
\draw[->] (d) -- (x2);
\draw[->] (d) -- (x3);
\end{tikzpicture}
    \caption{Causal DAG associated to our synthetic distribution.}
    \label{fig:graph_synthetic}
\end{figure}
\fi

The distribution is generated by the following set of structural equations:

\begin{align*}
    Y & \xleftarrow{} N_y; \\
    D & \xleftarrow{} N_d; \\
    G_1 & \xleftarrow{} Y + N_{G_1}; \\
    G_2 & \xleftarrow{} 2 \cdot Y + 2 \cdot D + N_{G_2}; \\
    G_3 & \xleftarrow{} D + N_{G_3}; \\
\end{align*}

\looseness -1 where $N_y$ and $N_d \sim Ber(0.5)$, $N_{G_i} \sim \mathcal{N}(0, 1)$. To create a dataset, we draw $1000$ samples from our synthetic distribution and use $200$ of them as test samples.

We then learn a representation that is invariant to $D$ and that is predictive towards $Y$ using our proposed loss. As a distributional distance, we use the \MMD loss with Gaussian kernel. The architecture of the encoder is a neural network with one hidden layer of size $10$ and a representation size of $5$. The hidden layer is followed by a batch normalization and a ReLU activation. We use a batch size of $64$ and train with the Adam optimizer \citep{kingma2014adam} for $200$ epochs, with a learning rate of $0.001$ and weight decay of $5 \times 10^{-5}$.

\begin{figure}[!ht]
    \vskip 0.2in
    \begin{center}
    %\begin{subfigure}
    \includegraphics[width=0.7\columnwidth]{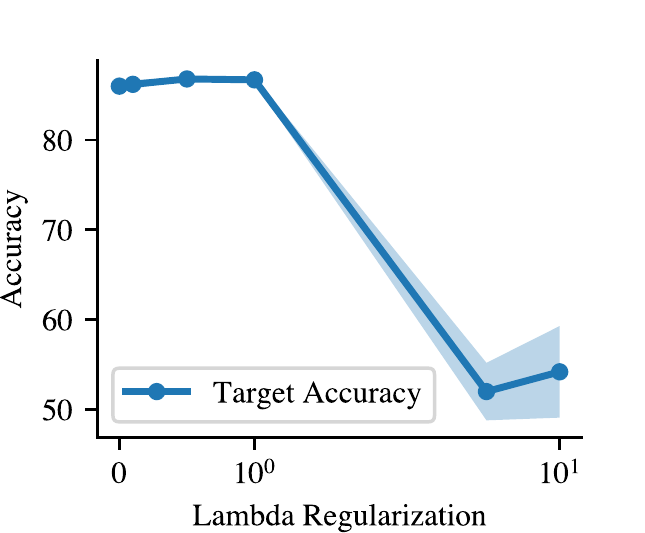}
    %\end{subfigure}
    %\begin{subfigure}
    \includegraphics[width=0.7\columnwidth]{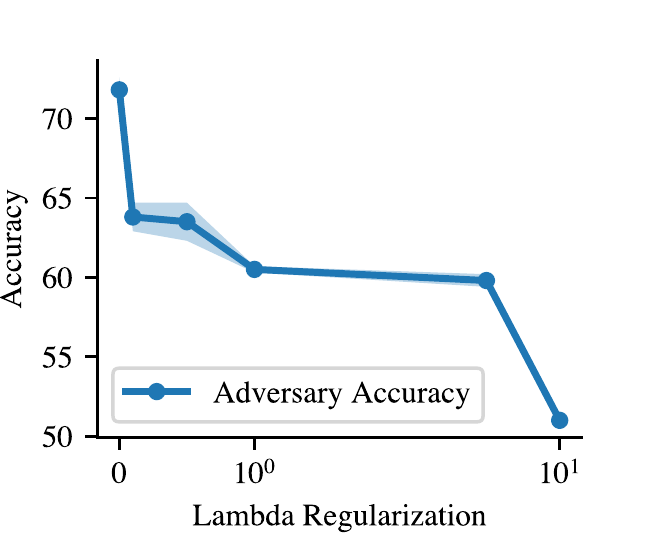}
    %\end{subfigure}
    \caption{Graphical visualization of our results on the synthetic dataset. We can observe the trade-off between invariance (Adversary Accuracy) and performance (Target Accuracy) for different values of $\lambda$. %The values here are the same than in \cref{tab:synthetic}.
    }
    \label{fig:synthetic}
    \end{center}
    \vskip -0.2in
\end{figure}

\looseness -1 After training the encoder, we freeze it and train two one layer linear discriminators: one to predict $Y$ and one to predict $D$. For each discriminator, we report the best achieved test accuracy. We run this experiment three times for each value of regularization $\lambda \in \{ 0.0, 0.1, 0.5, 1.0, 5.0, 10.0\}$. The results are summarized in \cref{fig:synthetic}.

As expected, we can observe a strong correlation between the strength of regularization and the strength of invariance. We achieved perfect invariance with $\lambda = 10.0$, where the adversary accuracy is $50\%$, but target accuracy is only $54.2\%$. This is expected: as $Y$ and $D$ are strongly correlated, removing information on $D$ in the representation also reduces the predictive power of the representation. There thus is a trade-off between performance and invariance that can be controlled via the value of $\lambda$. Finally, this experiment confirms that our proposed algorithm is a viable new method to enforce invariance.

\subsection{Fair Representation Learning experiments}
%\label{app:fair}

\looseness -1 We next present some experiments on fair representation learning. Here, we want to show that:
\begin{inparaenum}[(i)]
    \item Fair representation learning is also an invariant representation learning task, and it is covered by our unifying framework;
    \item Our algorithm is applicable to a wide range of tasks, as it also gives competitive results on this task;
    %\item We can control the strength of invariance via the hyperparameter $\lambda$;
    \item Fair representation learning datasets probably also follow our proposed data generation graph.
\end{inparaenum}

In the context of fair representation learning, the variable $D$ we want to be invariant to here corresponds to what is usually referred to as the \emph{sensitive} variable. A sensitive variable is a variable that should not have an effect on the predictions of a classifier or regressor. Some examples are the sex, the race or the age of an individual. If we can construct a representation that does not contain information about the sensitive variable, there is no way for a model built on top of this representation to base its prediction on the sensitive variable. Unfortunately, in many datasets, the sensitive variable is actually predictive for the target variable, i.e., the value we are trying to predict. This introduces a trade-off between fairness and accuracy of a model.

%\stefan{In the previous section you have a structure of dataset, propised models,  experiment design. You should have the same structure in all experiment settings then if possible. Compute seems not be to specified here?}

\paragraph{Datasets} We run experiments on two datasets from the UCI ML-repository \citep{asuncion2007uci}, the Adult and German dataset. The German dataset seeks to predict whether an individual has good or bad credit, while the sensitive attribute is the gender. The Adult dataset aims to predict whether the annual income of an individual is more or less than $50,000$\$, and the sensitive attribute is the gender. %See \cref{tab:fair_dataset} in the Appendix for a summary of the datasets. We also report the size of the majority class for the sensitive and target attribute for each dataset. 
A fair model should have a sensitive accuracy that is close or below the size of the majority sensitive class, while having a target accuracy as high as possible.

\paragraph{Experiment Design} To run our experiments, we reuse the code from \citet{roy2019mitigating} and add our model. We also empirically modify the default latent representation size such that it is optimizable using the \MMD distance. As for the synthetic experiment, after training the encoder, we freeze it and learn two discriminators: one for the target and one for the sensitive attribute. The target discriminators is trained for $100$ epochs and the adversary discriminator for $150$ epochs. We keep the best achieved test accuracy. The goal of this setup is to assess how much information can be extracted from the representation regarding the target and sensitive variables. %For each value of $\lambda$ regularization, we run the experiment three times, and report mean and standard deviation. We also report the best obtained model and compare it to other baselines. The goal is to show what performance we can potentially achieve with our algorithm and see how it compare with existing models.
In \cref{app:fair}, results for the German dataset as well as figures for both datasets representing the trade-off between performance and invariance can be found.

\iffalse
\begin{table}[H]
    \centering
    \caption{Main characteristics of the datasets used in our fair representation learning experiments. %\stefan{reference in main text but shift table to appendix}
    }
    \adjustbox{max width=\textwidth}{
    \begin{tabular}{lcccccc}
    \toprule
         Dataset &  Support of $D$ & Target Variable & Dataset Size & Input Size & Majority Sensitive & Majority Target \\
         \midrule
         Adult & \{ male, female \} & Income > $50,000$\$ & $45,222$ & $14$ & $67\%$ & $75\%$\\
         German & \{ male, female \} & Good or bad credit & $1,000$ & $20$ & $69\%$ & $71\%$\\
         %Extended Yale B & \{ upper right, lower right, lower left, upper left, front \} & 38 individuals & $1,286$ & $192 \times 168$ & $50\%$ & $2.6\%$\\
         \bottomrule
    \end{tabular}}
    
    \label{tab:fair_dataset}
\end{table}
\fi

\paragraph{Results (Adult Dataset)}

The encoder is a neural network with one hidden layer of size $7$, and a latent representation size of $2$. It is trained for $150$ epochs using the Adam optimizer \citep{kingma2014adam}, with learning rate  $1 \times 10^{-4}$ and weight decay  $5 \times 10^{-2}$. The discriminators are two-hidden-layer neural networks, with hidden layers of size $64$ and $32$. Both are optimized using Adam with learning rate of $0.001$ and weights decay of $0.001$. The learning rate of the discriminators is adjusted with Cosine Annealing. Train batch size is set to $128$ and test batch size to $1000$.

%Results are summarized in \cref{fig:fair}, as well as a comparison with other baselines in \cref{tab:adult_comp}. The results are as we expected: stronger regularization leads to stronger invariance towards the sensitive attribute. We also get that there is a trade-off between target accuracy and adversary accuracy, as the sensitive attribute is informative towards predicting the target. 
Results are summarized in \cref{tab:adult_comp}. Compared to other baselines, our best model performs well, as it has the best target accuracy for a slightly higher adversary accuracy. This shows that our method may offer a better trade-off as it allows for better performance for slightly lower invariance.

\begin{table*}[t]
\caption{Comparison to other existing models on the Adult dataset.%\stefan{where is coral and mmd?}
    } 
    \label{tab:adult_comp}
    \vskip 0.15in
\begin{center}
\begin{small}
\begin{sc}
       \begin{tabular}{lcc}
    \toprule
    Model & Target Accuracy & Adversary Accuracy\\
    \midrule
    \rowcolor{lightblue}
    \textbf{CausIRL with MMD (ours)} & \textbf{85.0} & 69.8\\
    ML-ARL \citep{xie2017controllable} & 84.4 & 67.7\\
    MaxEnt-ARL  \citep{roy2019mitigating}& 84.6 & 65.5\\
    LFR \citep{pmlr-v28-zemel13} & 82.3 & 67.0\\
    %VAE & & 66\\
    VFAE \citep{louizos2017variational} & 81.3 & 67.0\\
    Majority Classifier & 75.0 & 67.0\\
    \bottomrule
    \end{tabular}
    \end{sc}
\end{small}
\end{center}
\vskip -0.1in
\end{table*}

\subsection{Domain Generalization}
%We now test our algorithm on the task of \DG. In \DG, we are given a set of training domains, and we test the learned model on one unobserved domain. 
\paragraph{Datasets} For this experiment, we test on seven datasets: ColoredMNIST \citep{arjovsky2019invariant}, RotatedMNIST \citep{ghifary2015domain}, VLCS \citep{fang2013unbiased}, PACS \citep{Li_2017_ICCV}, OfficeHome \citep{venkateswara2017Deep}, TerraIncognita \citep{beery2018recognition} and DomainNet \citep{peng2019moment}. In the Appendix, \cref{tab:datasets} shows sample images for each dataset under different domains and \cref{tab:dg_datasets} presents each dataset's characteristics.

\paragraph{Experiment Design} We run our experiments with the DomainBed \citep{gulrajani2020search} testbed, which is a recent widely used testbed for \DG. We choose this setup as it allows for a highly fair and unbiased comparison with other existing models. DomainBed was designed to be reproducible, to give each algorithm the same amount of hyperparameter search, and to accurately estimate the variance in performance. Three model selection methods are considered: training-domain validation (all training models are pooled and a fraction of each of them is used as validation set), leave-one domain-out cross-validation (cross validation is performed using a different domain as validation, and the best models is retrained on all training domains) and test-domain validation set (a fraction of the test domain is used as validation set). The first two methods are closer to a realistic setting, whereas oracle validation allows us to evaluate whether there exists headroom for improvement. Training-domain validation assumes that all training domains and the test domain follow a similar distribution, as we pool all the training domains during training. On the other hand, leave-one domain-out cross-validation is closer to our assumption, as it optimizes for generalization to an unseen domain that is assumed to follow a different distribution.

\paragraph{Proposed Models} \looseness -1 We take two existing models, MMD and CORAL, based on matching distributions across domains, and propose two new models, CausIRL with MMD and with CORAL. These two new algorithms simply consist in changing how the regularization loss is computed according to our proposed algorithm, i.e., instead of taking pairwise distances across domains, we compute distances between batches that follow different domain distributions. We thus want to see if this simple change in the algorithm leads to better performance, which may be, as we conjecture, due to a better trade-off between performance and invariance, as well as the fact that it may be easier to optimize, especially in the presence of many domains. The hyperparameter $\lambda$ is drawn randomly in $10^{\text{Uniform}(-1, 1)}$. In the following, we present the results for two model selection methods, while the results for the last one (training-domain validation) can be found in \cref{app:dg}. Tables with comparisons to a larger number of baselines can also be found in the appendix.

\subsubsection{Model Selection: Leave-One-Domain-Out Cross-Validation}

We now look at the \DG experiment results for the leave-one-domain-out cross-validation model selection method. The results are summarized in \cref{tab:dg_leave}. %The result for this model selection method are the most relevant as it closely follows our assumptions on the training and test distributions. . 

Here, the overall performance of CausIRL with CORAL is almost identical to CORAL. Nevertheless, there are some differences when looking at the performance on individual datasets. CausIRL with CORAL overperform CORAL on PACS, TerraIncognita and DomainNet, whereas CORAL performs better on VLCS and OfficeHome. However we should note that only the overperformance of CausIRL with CORAL over CORAL on DomainNet is statistically significant when looking at the confidence intervals of the average accuracies.  

For CausIRL with MMD, we observe a significant boost in the overall performance compared to MMD. CausIRL with MMD performs better on almost all datasets, and we also observe a significant leap in performance on DomainNet going from $23.4\%$ to $38.9\%$. %This shows that the inductive bias of our algorithm matches the real inductive bias of \DG far more accurately.

\begin{table*}[t]
\centering
\caption{Domain Generalization experimental results for the leave-one-domain-out cross-validation model selection method.}
\label{tab:dg_leave}
\vskip 0.15in
\begin{center}
\begin{tiny}
\begin{sc}
%\adjustbox{max width=\pdfpagewidth}{%
\begin{tabular}{lcccccccc}
\toprule
\textbf{Algorithm}        & \textbf{ColoredMNIST}     & \textbf{RotatedMNIST}     & \textbf{VLCS}             & \textbf{PACS}             & \textbf{OfficeHome}       & \textbf{TerraIncognita}   & \textbf{DomainNet}        & \textbf{Avg}              \\
\midrule
\rowcolor{lightblue}
\textbf{CausIRL with CORAL (ours)}                 & 39.1 $\pm$ 2.0            & 97.8 $\pm$ 0.1            & 76.5 $\pm$ 1.0            & 83.6 $\pm$ 1.2            & 68.1 $\pm$ 0.3            & 47.4 $\pm$ 0.5            & \textbf{41.8 $\pm$ 0.1}            & 64.9                      \\
CORAL                     & 39.7 $\pm$ 2.8            & 97.8 $\pm$ 0.1            & \textbf{78.7} $\pm$ 0.4            & 82.6 $\pm$ 0.5            & \textbf{68.5} $\pm$ 0.2            & 46.3 $\pm$ 1.7            & 41.1 $\pm$ 0.1            & \textbf{65.0}                      \\
\rowcolor{lightblue}
\textbf{CausIRL with MMD (ours)}                   & 36.9 $\pm$ 0.2            & 97.6 $\pm$ 0.1            & 78.2 $\pm$ 0.9            & \textbf{84.0} $\pm$ 0.9            & 65.1 $\pm$ 0.7            & \textbf{47.9} $\pm$ 0.3            & 38.9 $\pm$ 0.8            & 64.1                      \\
MMD                       & 36.8 $\pm$ 0.1            & 97.8 $\pm$ 0.1            & 77.3 $\pm$ 0.5            & 83.2 $\pm$ 0.2            & 60.2 $\pm$ 5.2            & 46.5 $\pm$ 1.5            & 23.4 $\pm$ 9.5            & 60.7                      \\
\midrule
ERM                       & 36.7 $\pm$ 0.1            & 97.7 $\pm$ 0.0            & 77.2 $\pm$ 0.4            & 83.0 $\pm$ 0.7            & 65.7 $\pm$ 0.5            & 41.4 $\pm$ 1.4            & 40.6 $\pm$ 0.2            & 63.2                      \\
IRM                       & 40.3 $\pm$ 4.2            & 97.0 $\pm$ 0.2            & 76.3 $\pm$ 0.6            & 81.5 $\pm$ 0.8            & 64.3 $\pm$ 1.5            & 41.2 $\pm$ 3.6            & 33.5 $\pm$ 3.0            & 62.0                      \\
GroupDRO                  & 36.8 $\pm$ 0.1            & 97.6 $\pm$ 0.1            & 77.9 $\pm$ 0.5            & 83.5 $\pm$ 0.2            & 65.2 $\pm$ 0.2            & 44.9 $\pm$ 1.4            & 33.0 $\pm$ 0.3            & 62.7                      \\
DANN                      & \textbf{40.7} $\pm$ 2.3            & 97.6 $\pm$ 0.2            & 76.9 $\pm$ 0.4            & 81.0 $\pm$ 1.1            & 64.9 $\pm$ 1.2            & 44.4 $\pm$ 1.1            & 38.2 $\pm$ 0.2            & 63.4                      \\
CDANN                     & 39.1 $\pm$ 4.4            & 97.5 $\pm$ 0.2            & 77.5 $\pm$ 0.2            & 78.8 $\pm$ 2.2            & 64.3 $\pm$ 1.7            & 39.9 $\pm$ 3.2            & 38.0 $\pm$ 0.1            & 62.2                      \\
VREx                      & 36.9 $\pm$ 0.3            & 93.6 $\pm$ 3.4            & 76.7 $\pm$ 1.0            & 81.3 $\pm$ 0.9            & 64.9 $\pm$ 1.3            & 37.3 $\pm$ 3.0            & 33.4 $\pm$ 3.1            & 60.6                      \\
\bottomrule
\end{tabular}
%}
\end{sc}
\end{tiny}
\end{center}
\vskip -0.1in    
\end{table*}

\subsubsection{Model Selection: Test-Domain Validation Set (Oracle)}

Finally, we here look at the \DG experiment results for the test-domain validation set model selection method. The results are summarized in \cref{tab:dg_oracle}. This setting is less realistic as we have access to test samples during training, but it is still useful as it shows the best possible model for each algorithm. It allows us to evaluate whether there is headroom for improvement for each algorithm and to see which algorithm has the inductive bias that more closely fit the task. 

For both CausIRL with CORAL and CausIRL with MMD, we observe a better overall performance compared to their vanilla counterparts. We even have that CausIRL with CORAL is the best overall performing algorithm among the evaluated algorithms. Once again, we observe a large difference in performance on DomainNet between MMD and CausIRL with MMD, going from an average accuracy of $23.5\%$ to $40.6\%$. We also again have that CausIRL with CORAL is the best algorithm for DomainNet compared to all the other algorithms.

\begin{table*}[]
\centering
\caption{Domain Generalization experimental results for the test-domain validation set model selection method.}
\label{tab:dg_oracle}
\vskip 0.15in
\begin{center}
\begin{tiny}
\begin{sc}
%\adjustbox{max width=\textwidth}{%
\begin{tabular}{lcccccccc}
\toprule
\textbf{Algorithm}        & \textbf{ColoredMNIST}     & \textbf{RotatedMNIST}     & \textbf{VLCS}             & \textbf{PACS}             & \textbf{OfficeHome}       & \textbf{TerraIncognita}   & \textbf{DomainNet}        & \textbf{Avg}              \\
\midrule
\rowcolor{lightblue}
\textbf{CausIRL with CORAL (ours)}                & 58.4 $\pm$ 0.3            & 98.0 $\pm$ 0.1            & 78.2 $\pm$ 0.1            & \textbf{87.6 $\pm$ 0.1}            & 67.7 $\pm$ 0.2            & \textbf{53.4} $\pm$ 0.4            & \textbf{42.1 $\pm$ 0.1}            & \textbf{69.4}                      \\
CORAL                     & 58.6 $\pm$ 0.5            & 98.0 $\pm$ 0.0            & 77.7 $\pm$ 0.2            & 87.1 $\pm$ 0.5            & \textbf{68.4 $\pm$ 0.2}            & 52.8 $\pm$ 0.2            & 41.8 $\pm$ 0.1            & 69.2                      \\
\rowcolor{lightblue}
\textbf{CausIRL with MMD (ours)}                  & 63.7 $\pm$ 0.8            & 97.9 $\pm$ 0.1            & 78.1 $\pm$ 0.1            & 86.6 $\pm$ 0.7            & 65.2 $\pm$ 0.6            & 52.2 $\pm$ 0.3            & 40.6 $\pm$ 0.2            & 69.2                      \\
MMD                       & 63.3 $\pm$ 1.3            & 98.0 $\pm$ 0.1            & 77.9 $\pm$ 0.1            & 87.2 $\pm$ 0.1            & 66.2 $\pm$ 0.3            & 52.0 $\pm$ 0.4            & 23.5 $\pm$ 9.4            & 66.9                      \\
\midrule
ERM                       & 57.8 $\pm$ 0.2            & 97.8 $\pm$ 0.1            & 77.6 $\pm$ 0.3            & 86.7 $\pm$ 0.3            & 66.4 $\pm$ 0.5            & 53.0 $\pm$ 0.3            & 41.3 $\pm$ 0.1            & 68.7                      \\
IRM                       & \textbf{67.7} $\pm$ 1.2            & 97.5 $\pm$ 0.2            & 76.9 $\pm$ 0.6            & 84.5 $\pm$ 1.1            & 63.0 $\pm$ 2.7            & 50.5 $\pm$ 0.7            & 28.0 $\pm$ 5.1            & 66.9                      \\
GroupDRO                  & 61.1 $\pm$ 0.9            & 97.9 $\pm$ 0.1            & 77.4 $\pm$ 0.5            & 87.1 $\pm$ 0.1            & 66.2 $\pm$ 0.6            & 52.4 $\pm$ 0.1            & 33.4 $\pm$ 0.3            & 67.9                      \\
DANN                      & 57.0 $\pm$ 1.0            & 97.9 $\pm$ 0.1            & 79.7 $\pm$ 0.5            & 85.2 $\pm$ 0.2            & 65.3 $\pm$ 0.8            & 50.6 $\pm$ 0.4            & 38.3 $\pm$ 0.1            & 67.7                      \\
CDANN                     & 59.5 $\pm$ 2.0            & 97.9 $\pm$ 0.0            & \textbf{79.9} $\pm$ 0.2            & 85.8 $\pm$ 0.8            & 65.3 $\pm$ 0.5            & 50.8 $\pm$ 0.6            & 38.5 $\pm$ 0.2            & 68.2                      \\
VREx                      & 67.0 $\pm$ 1.3            & 97.9 $\pm$ 0.1            & 78.1 $\pm$ 0.2            & 87.2 $\pm$ 0.6            & 65.7 $\pm$ 0.3            & 51.4 $\pm$ 0.5            & 30.1 $\pm$ 3.7            & 68.2                      \\
\bottomrule
\end{tabular}
%}
\end{sc}
\end{tiny}
\end{center}
\vskip -0.1in     
\end{table*}

\subsection{Real-World Domain Generalization}

\begin{table}[ht]%{l}{0.5\textwidth}
\centering
\caption{Performance results of our proposed models on Camelyon17 and RxRx1 compared to other baselines.}
\label{tab:results_real}
\vskip 0.15in
\begin{center}
\begin{tiny}
\begin{sc}
%\adjustbox{width=\columnwidth}{
\begin{tabular}{lcc}
\toprule
\textbf{Algorithm}        & \textbf{Camelyon17}     & 
\textbf{RxRx1}              \\
\midrule
\rowcolor{lightblue}
\textbf{CausIRL with CORAL (ours)}   & 62.7 $\pm$ 9.4  & 29.0 $\pm$ 0.2\\
CORAL      & 59.5 $\pm$ 7.7   & 28.4 $\pm$ 0.3\\
\rowcolor{lightblue}
\textbf{CausIRL with MMD (ours)}  & 63.4 $\pm$ 11.2  & 28.9 $\pm$ 0.1 \\
MMD & 64.6 $\pm$ 10.5 & 28.2 $\pm$ 0.2\\
ERM & \textbf{70.3} $\pm$ 6.4 & \textbf{29.9} $\pm$ 0.4\\
GroupDRO &  68.4 $\pm$ 7.3 & 23.0 $\pm$ 0.3\\
IRM & 64.2 $\pm$ 8.1 & 9.9 $\pm$ 1.4\\
\bottomrule
\end{tabular}
%}
\end{sc}
\end{tiny}
\end{center}
\vskip -0.1in

\end{table}

\looseness -1 In this section, we run experiments on more realistic distributional shifts. We use the Wilds \citep{wilds2021} benchmark and run experiments on two datasets: Camelyon17 \citep{bandi2018detection} and RxRx1 \citep{taylor2019rxrx1}. Camelyon17 consists in predicting whether a region of tissue contains tumor tissue, while being invariant to the hospitals where the sample was taken. The goal is to obtain a model that generalizes across hospitals, as hospital specific artifacts of the data collection process can vary. RxRx1 consists of cell images, where the cells received some genetic treatment (as well as no treatment). The goal is to predict the genetic treatment among $1,139$ possible treatments. Here, we want to be invariant to the \emph{batch} the cells come from, as it is a common observation that batch effects can greatly alter the results. 

We test our two proposed models, CausIRL with CORAL and with MMD on both datasets. For the RxRx1 dataset, we use the same hyperparameters than for the CORAL model in the Wilds implementation. For Camelyon17, we change the number of group per batch to three and the batch size to $60$. The results are summarized in \cref{tab:results_real}. As for the \DG experiments on DomainBed before, we observe that CausIRL with CORAL performs better than CORAL. Moreover, CausIRL with MMD performs slightly better than CausIRL with CORAL on Camelyon17 and similarly on RxRx1. Unfortunately, all models perform worse than simple ERM. Indeed, real world datasets exhibit far more complex data generating processes, which makes finding suitables heuristics highly difficult. Nevertheless, we again observe that our proposed models work competitively even on a realistic dataset, and that our proposed algorithm to compute the distributional distance regularization is better than how it is traditionally done.

\vspace{-0.25em}
\section{Conclusion and Future Work}

\looseness -1 In this work, we provided a causal perspective on invariant representation learning. %We took a causal perspective to define what is invariance. 
%We proposed a causal \DAG to unify different machine learning task under the same framework and cast them as a invariant representation learning tasks. 
%We defined style variables in the context of our framework and developed theory on what conditions are necessary or sufficient to be invariant towards the style variables. 
Based on this causal perspective and the assumptions on the data generating process, we then proposed a new, simple and versatile algorithm for enforcing invariance to $D$ in the learned representations. As our regularization is softer than traditional methods, we argue that it offers a better trade-off between performance and invariance, which is supported by our empirical results. Furthermore, as our method is simple and non task specific, it should be widely applicable. As it is easily implementable, it can be a viable additional option for practitioners. %based on our data generation assumptions. %to enforce invariance, with the conjectured property that is enforces stronger invariance towards the style variables.
Lastly, we empirically demonstrated that our algorithm is versatile as it works on a diverse set of tasks and datasets. In particular, it performs strongly in \DG, where we obtain state-of-the-art performance. %We observe that it may have interesting optimization properties, especially when the support of the sensitive variable $D$ is large.

%However, there still exists some interesting directions to follow up on the findings of this work. Particularly, we didn't study the case where the variable $D$ is multi-dimensional or continuous. Also, it would be interesting to study the case where $D$ has on effect on $Y$, and how to be invariant to $D$ while keeping information about $Y$. This is a settings that is often observed and which requires a modification of our proposed algorithm.

\newpage

%\section{Reproducibility Statement}

%Regarding the theoretical results, all results are based on the assumptions of \Cref{fig:graph}, and the proofs can be found in \Cref{app:proofs}. For the fair representation learning experiments, the architectures and training procedures are precisely described in \Cref{ssec:fair}, and we reused the code of \citet{roy2019mitigating} to ensure a fair comparison to existing models. For the \DG experiments, we used two benchmarks, DomainBed \citep{gulrajani2020search} and Wilds \citep{wilds2021}, which were designed to be reproducible and unbiased. The hyperparameter used can be found in the main text. Regarding the implementation of our algorithm, the pseudo-code in \cref{alg:invariance} and the python code snippets in the Appendix should be sufficient to accurately implement our proposed models. For all experiments, our implementation of the MMD and CORAL distance are taken from the DomainBed \citep{gulrajani2020search} code. Our code implementations are provided with all the necessary details in the corresponding sections in the appendix. For the implementation details for fair representation learning we refer to the detailed documentation of the code in section \ref{app:fair} and for domain generalization we refer to section \ref{app:dg}. 


\bibliography{main}
\bibliographystyle{icml2022}

%%%%%%%%%%%%%%%%%%%%%%%%%%%%%%%%%%%%%%%%%%%%%%%%%%%%%%%%%%%%%%%%%%%%%%%%%%%%%%%
%%%%%%%%%%%%%%%%%%%%%%%%%%%%%%%%%%%%%%%%%%%%%%%%%%%%%%%%%%%%%%%%%%%%%%%%%%%%%%%
% APPENDIX
%%%%%%%%%%%%%%%%%%%%%%%%%%%%%%%%%%%%%%%%%%%%%%%%%%%%%%%%%%%%%%%%%%%%%%%%%%%%%%%
%%%%%%%%%%%%%%%%%%%%%%%%%%%%%%%%%%%%%%%%%%%%%%%%%%%%%%%%%%%%%%%%%%%%%%%%%%%%%%%
\newpage
\appendix
\onecolumn

\section{Additional Background}
\label{app:back}

\subsection{Distributional Distances}

The main goal of this work is to study how invariance can be enforced by regularizing different latent spaces to have the same distribution. To this end, we thus need a differentiable distance or divergence between distributions that can be minimized during training. We here present the most commonly used distances in the literature. 

\subsubsection{Adversarial}

Adversarial training was first introduced in \cite{Goodfellow2014} as a new method for Generative modeling. Based on game theory, it can intuitively be described as a two player game, where each player is parameterized by a neural network. The Generator is a function that maps its input distribution to an output distribution. We call it the generated distribution and denote it by $p_g$. On the other hand, a Discriminator tries to distinguish between samples coming from the target dataset and samples produced by the Generator. At convergence, the Generator produces data that is distributed similarly to the target distribution, and thus it becomes impossible for the Discriminator to distinguish samples.

Formally, the objective of the two-player minimax game reads:

\begin{align}
\label{eq:gan}
    \min_{G} \max_{D} V(D, G) & = \mathbb{E}_{\mathbf{x} \sim p_{data}(\mathbf{x})} \left[\log D(\mathbf{x})\right]  + \mathbb{E}_{\mathbf{z} \sim p_{\mathbf{z}}(\mathbf{z})} \left[\log \left( 1 - D (G(\mathbf{z})) \right)\right]
\end{align}

where $\mathbf{z}$ is the input, $\mathbf{x}$ comes from the target distributions, and the Discriminator $D$ should output $1$ when its input is a samples from the target, and $0$ otherwise. If the Discriminator is optimal for a given $G$, \cref{eq:gan} can be rewritten to show that the Generator actually minimizes the \JSD between the generated and target distribution. 
\begin{align*}
    JSD\left(P  \middle \|  Q\right) = \frac{1}{2} \KL\left ( P  \middle \|  \frac{1}{2} \left(P + Q\right) \right ) + 
    \frac{1}{2} \KL\left (Q  \middle \|  \frac{1}{2} \left(P + Q\right) \right ),
\end{align*}
where $\KL$ is the \KLd divergence.
It also can be shown that if both networks have sufficient capacity, and if the Discriminator is trained to optimality after each optimization step of the Generator, then the distribution of the Generator converges to the target distribution.

Adversarial training can thus be seen as a proxy distributional distance, which corresponds to the \JSD at convergence. This concept of adversarial training has been extended to be used as a regularizer for latent spaces. It can for example be used to enforce a prior distribution on the latent space \citep{makhzani2015adversarial}. It can also be used to enforce two latent spaces to have the same distribution. Its use is often justified as wanting two latent spaces to seem \emph{indistinguishable} for an adversary, which is supposed to force the encoder to discard what is not constant across the two input distribution. We argue that adversarial training is theoretically equivalent to minimizing any distributional divergence, and that only their optimization properties differentiate them. We will also later clarify the intuition of trying to discard the \emph{idiosyncratic in favor of the universal}, and what it actually corresponds to when we look at the data generation process of a given dataset.

\subsubsection{Maximum Mean Discrepancy}

\MMD \citep{gretton2006kernel} is a distance based on empirical samples from two distributions, based on the distance between the means of the two sets of samples mapped into a \RKHS. Let $\{ X \} \sim P$ and $\{ X' \} \sim Q$. Then, we have:

\begin{align*}
    MMD(X, X')^2 & = \left \| \frac{1}{n} \sum_{i = 1}^n \phi(x_i) - \frac{1}{n'} \sum_{i = 1}^{n'} \phi(x'_i) \right \|\\
    & = \frac{1}{n^2} \sum_{i, j = 1}^n k(x_i, x_j) + \frac{1}{{n'}^2} \sum_{i, j = 1}^{n'} k(x'_i, x'_j)  - \frac{2}{n \cdot n'} \sum_{i = 1}^n \sum_{j = 1}^{n'} k(x_i, x'_j) ,\\
\end{align*}

where $k(\cdot, \cdot)$ is the associated kernel. One commonly used kernel is the Gaussian kernel $k(x, x') = e^{-\lambda \| x - x' \|^2}$. Asymptotically, for a universal kernel such as the Gaussian kernel, $MMD(X, X') = 0$ if and only if $P = Q$. Minimizing the \MMD distance during training can thus be used to align two distributions.

%\section{Proofs of theorems}
%\label{app:proofs}

\iffalse
\thmequivalence*
\begin{proof}
$\implies$ As $z$ is a descendant of $d$, the mechanism $p(z|d)$ is invariant to the distribution of $d$.
    \begin{align*}
         p^{do(d = N_d)}(z) & = \int p^{do(d = N_d)}(z, d')  \mathrm{d}d' 
        = \int p^{do(d = N_d)}(z|d') N_d (d') \mathrm{d}d' 
        & = p(z) \int N_d (d') \mathrm{d}d' \\& = p(z)
    \end{align*}
 $\impliedby$ As $p(z)$ is constant for all distribution of $d$, then it is also constant for deterministic distributions, i.e $\delta_d$. 
    \begin{align*}
         p(z) &  = \int p(z|d') \delta_d (d') \mathrm{d}d'= p(z|d)
    \end{align*}
    This holds for all values of $d$, which implies that $I(Z;D) = 0$.
\end{proof}

\thmnecessary*
\begin{proof}
    \begin{align*}
        p(z | d = 1) - p(z | d = 2) & = 
        \int p(z | S, d = 1) [p(S | d = 1) dS - \int p(z | S, d = 2) p(S | d = 2)]  dS\\
        & = \int p(z | S) [p(S | d = 1) - p(S | d = 2)]  dS\\
        & = p(z) \int [p(S | d = 1) - p(S | d = 2)] dS\\
        & = 0
    \end{align*}
\end{proof}

\thmsufficient*
\begin{proof}
    From the definition of total causal effect, let's suppose by contradiction that there exists an intervention on $S$ such that $p^{do(S = \tilde{N}_s)}(z) \neq p(z)$. Let $\tilde{d}$ denote the domain that correspond to this intervention. We then have a value of $D$ such that $p(z|\tilde{d}) \neq p(z)$, which is a contradiction to $Z$ being independent to $D$.
\end{proof}
\fi

\section{Fair representation learning supplements}
\label{app:fair}

\begin{figure}[ht]
\vskip 0.2in
\begin{center}
    %\begin{subfigure}
    \begin{minipage}{.4\columnwidth}
    \includegraphics[width=\columnwidth]{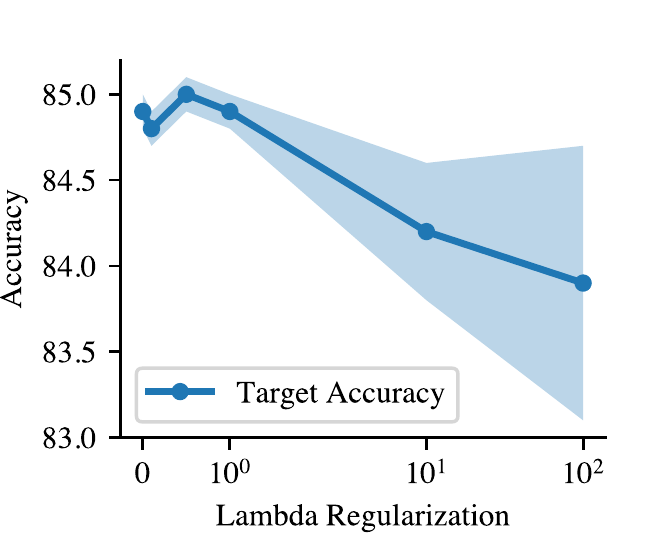}
    %\end{subfigure}
    %\begin{subfigure}
    \end{minipage}
    \begin{minipage}{.4\columnwidth}
    \includegraphics[width=\columnwidth]{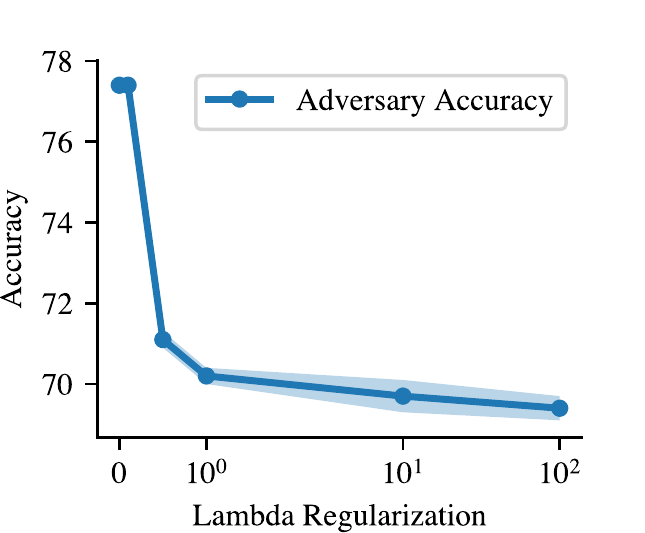}
    \end{minipage}
    \caption{Graphical visualization of our results on the Adult dataset. We can observe the trade-off between invariance (Adversary Accuracy) and performance (Target Accuracy) for different values of $\lambda$. }
    %\end{subfigure}
    %\begin{subfigure}
    \label{fig:fair_adult}
\end{center}
\vskip -0.2in
\end{figure}

\begin{figure}[ht]
\vskip 0.2in
\begin{center}   
    \begin{minipage}{.4\columnwidth}
    \includegraphics[width=\columnwidth]{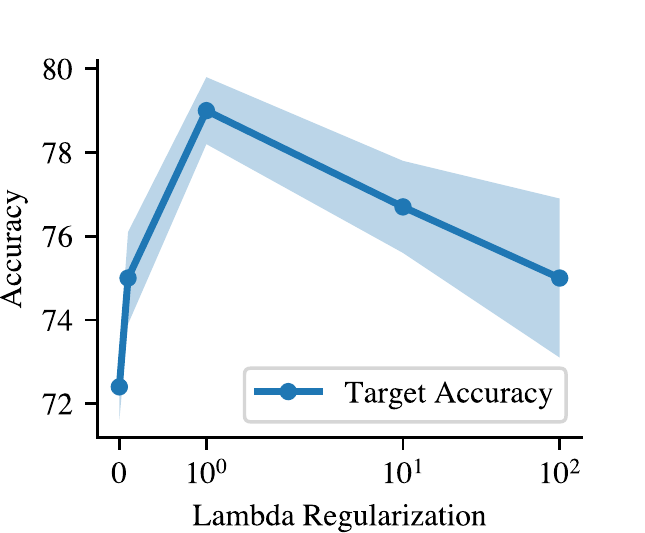}
    %\end{subfigure}
    %\begin{subfigure}
    \end{minipage}
    \begin{minipage}{.4\columnwidth}
    \includegraphics[width=\columnwidth]{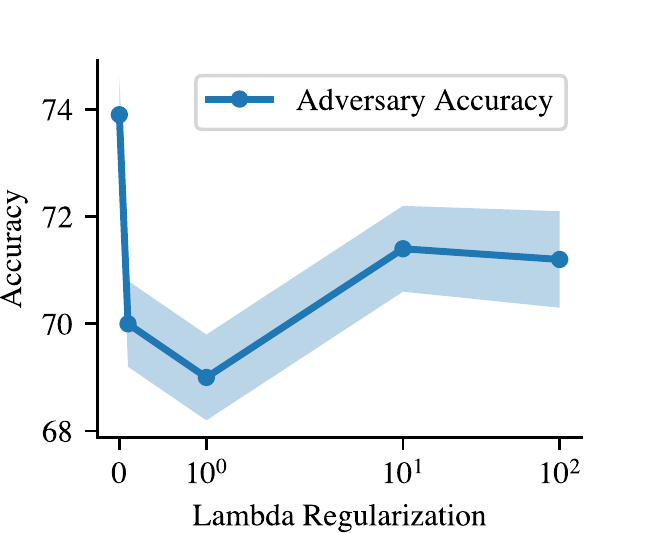}
    %\end{subfigure}
    \end{minipage}
    \caption{Graphical visualization of our results on the German dataset. We can observe the trade-off between invariance (Adversary Accuracy) and performance (Target Accuracy) for different values of $\lambda$.}
    \label{fig:fair_ger}
\end{center}
\vskip -0.2in
\end{figure}

\paragraph{German Dataset}

The encoder is a neural network with two hidden layers of size $15$ and $8$, and a latent representation size of $32$. It is trained for $150$ epochs using the Adam optimizer, with learning rate of $1 \times 10^{-4}$ and weight decay of $5 \times 10^{-2}$. The discriminators are two-hidden-layer neural networks, with hidden layers of size $10$. Both are optimized using Adam with learning rate of $0.001$ and weights decay of $0.001$. The learning rate of the discriminators is adjusted with Cosine Annealing. Train batch size is set to $64$ and test batch size to $100$.

Results are summarized in \cref{fig:fair_ger}, as well as a comparison with other baselines in \cref{tab:german_comp}. Here, we observe that $1.0$ is a clear optimal value for $\lambda$, as it gives the highest target accuracy and the lowest adversary accuracy. A bit more surprisingly, we observe that higher regularization can give lesser invariance, which we can interpret as a form of over-regularization. Compared to other methods, we obtain competitive results as we get the smallest adversary accuracy, even below the majority prediction, while still obtaining the second best target accuracy.

\begin{table*}[t]
\caption{Comparison to other existing models on the German dataset. %\stefan{why is there no CausIRL with CORAL and why no MMD?}
    }
    \label{tab:german_comp}
    \vskip 0.15in
\begin{center}
\begin{small}
\begin{sc}
    
    \begin{tabular}{lcc}
    \toprule
    Model & Target Accuracy & Adversary Accuracy\\
    \midrule
    \rowcolor{lightblue}
    \textbf{CausIRL with MMD (ours)} & 80.3 & 67.0\\
    ML-ARL \citep{xie2017controllable} & 74.4 & 80.2\\
    MaxEnt-ARL \citep{roy2019mitigating} & 86.3 & 72.7\\
    LFR \citep{pmlr-v28-zemel13} & 72.3 & 80.5\\
    %VAE & 72.5 & 79.5\\
    VFAE \citep{louizos2017variational} & 72.7 & 79.7\\
    Majority Classifier & 71.0 & 69.0\\
    \bottomrule
    \end{tabular}
\end{sc}
\end{small}
\end{center}
\vskip -0.1in
\end{table*}

\paragraph{Compute Resources} We run the experiements on \textsc{NVIDIA GeForce RTX 2080 Ti} GPUs.

\section{\DG supplements}
\label{app:dg}

\begin{table}
    \caption{Datasets used in our \DG experiments, with sample images for each of them. This table is taken from \cite{gulrajani2020search}.}
    \begin{center}
    \begin{tabular}{lcccccc}
        \toprule
        \textbf{Dataset} & \multicolumn{6}{l}{\textbf{Domains}} \\
        \midrule
        & \tiny{+90\%} & \tiny{+80\%} & \tiny{-90\%} & & & \\
        Colored MNIST &
            \raisebox{-.5\height}{\includegraphics[width=25pt, height=25pt]{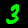}} &
            \raisebox{-.5\height}{\includegraphics[width=25pt, height=25pt]{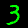}} &
            \raisebox{-.5\height}{\includegraphics[width=25pt, height=25pt]{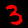}} & & &
            \\
        & \multicolumn{6}{l}{\tiny{\emph{(degree of correlation between color and label)}}} \\
        & \tiny{0$^{\circ}$} & \tiny{15$^{\circ}$} & \tiny{30$^{\circ}$} & \tiny{45$^{\circ}$} & \tiny{60$^{\circ}$} & \tiny{75$^{\circ}$} \\
        Rotated MNIST &
            \raisebox{-.5\height}{\includegraphics[width=25pt, height=25pt]{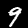}} &
            \raisebox{-.5\height}{\includegraphics[width=25pt, height=25pt]{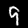}} &
            \raisebox{-.5\height}{\includegraphics[width=25pt, height=25pt]{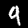}} &
            \raisebox{-.5\height}{\includegraphics[width=25pt, height=25pt]{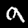}} &
            \raisebox{-.5\height}{\includegraphics[width=25pt, height=25pt]{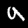}} &
            \raisebox{-.5\height}{\includegraphics[width=25pt, height=25pt]{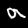}}
            \\
        & \tiny{Caltech101} & \tiny{LabelMe} & \tiny{SUN09} & \tiny{VOC2007} & & \\
        VLCS &
            \raisebox{-.5\height}{\includegraphics[width=25pt, height=25pt]{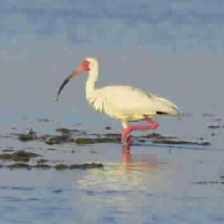}} &
            \raisebox{-.5\height}{\includegraphics[width=25pt, height=25pt]{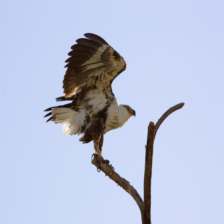}} &
            \raisebox{-.5\height}{\includegraphics[width=25pt, height=25pt]{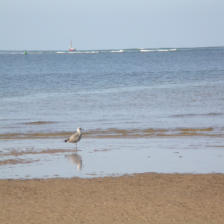}} &
            \raisebox{-.5\height}{\includegraphics[width=25pt, height=25pt]{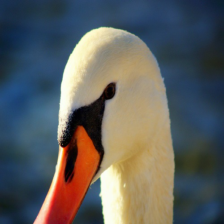}} & &
            \\
        & \tiny{Art} & \tiny{Cartoon} & \tiny{Photo} & \tiny{Sketch} & & \\
        PACS &
            \raisebox{-.5\height}{\includegraphics[width=25pt, height=25pt]{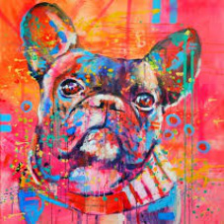}} &
            \raisebox{-.5\height}{\includegraphics[width=25pt, height=25pt]{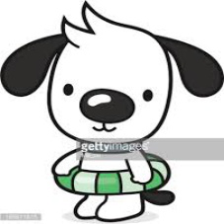}} &
            \raisebox{-.5\height}{\includegraphics[width=25pt, height=25pt]{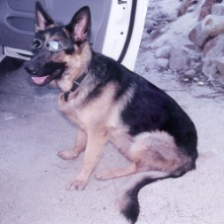}} &
            \raisebox{-.5\height}{\includegraphics[width=25pt, height=25pt]{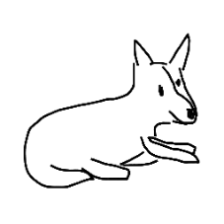}} & &
            \\
        & \tiny{Art} & \tiny{Clipart} & \tiny{Product} & \tiny{Photo} & & \\
        Office-Home &
            \raisebox{-.5\height}{\includegraphics[width=25pt, height=25pt]{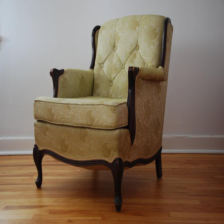}} &
            \raisebox{-.5\height}{\includegraphics[width=25pt, height=25pt]{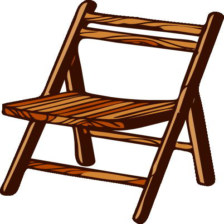}} &
            \raisebox{-.5\height}{\includegraphics[width=25pt, height=25pt]{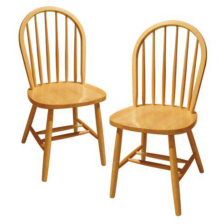}} &
            \raisebox{-.5\height}{\includegraphics[width=25pt, height=25pt]{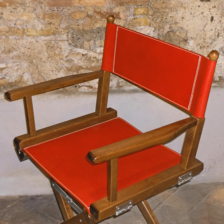}} & &
            \\
        & \tiny{L100} & \tiny{L38} & \tiny{L43} & \tiny{L46} & & \\
        Terra Incognita &
            \raisebox{-.5\height}{\includegraphics[width=25pt, height=25pt]{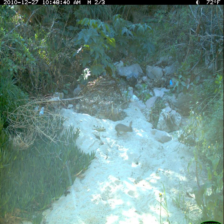}} &
            \raisebox{-.5\height}{\includegraphics[width=25pt, height=25pt]{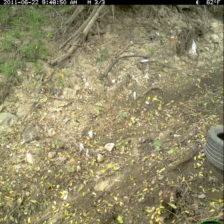}} &
            \raisebox{-.5\height}{\includegraphics[width=25pt, height=25pt]{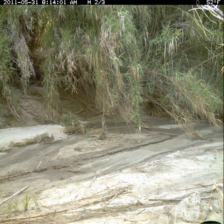}} &
            \raisebox{-.5\height}{\includegraphics[width=25pt, height=25pt]{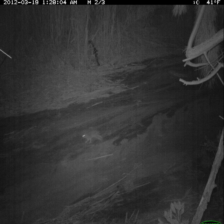}} &
            \\
        & \multicolumn{6}{l}{\tiny{\emph{(camera trap location)}}} \\
        & \tiny{Clipart} & \tiny{Infographic} & \tiny{Painting} & \tiny{QuickDraw} & \tiny{Photo} & \tiny{Sketch} \\
        DomainNet &
            \raisebox{-.5\height}{\includegraphics[width=25pt, height=25pt]{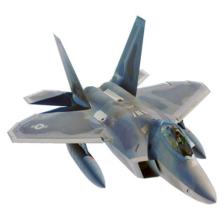}} &
            \raisebox{-.5\height}{\includegraphics[width=25pt, height=25pt]{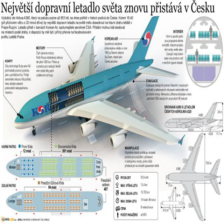}} &
            \raisebox{-.5\height}{\includegraphics[width=25pt, height=25pt]{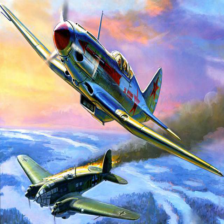}} &
            \raisebox{-.5\height}{\includegraphics[width=25pt, height=25pt]{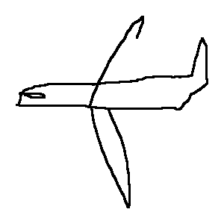}} &
            \raisebox{-.5\height}{\includegraphics[width=25pt, height=25pt]{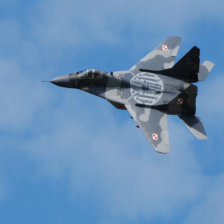}} &
            \raisebox{-.5\height}{\includegraphics[width=25pt, height=25pt]{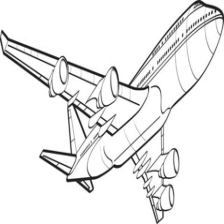}}
            \\
            \bottomrule
    \end{tabular}
    
    \label{tab:datasets}
    \end{center}
\end{table}

\begin{table}[]
    \centering
    \caption{Description of the datasets used in our \DG experiments}
    \adjustbox{max width=\columnwidth}{
    \begin{tabular}{lcccc}
        \toprule
         \textbf{Dataset Name} & \textbf{Support of $D$} & \textbf{Number of Samples} & \textbf{Image Dimensions} & \textbf{Number of Classes} \\
         \midrule
         ColoredMNIST \citep{arjovsky2019invariant} & $\{0.1, 0.3, 0.9\}$ & $70,000$ & $(2, 28, 28)$ & $2$ \\
         RotatedMNIST \citep{ghifary2015domain} & $\{ 0, 15, 30, 45, 60, 75 \}$ &  $70,000$ & $(1, 28, 28)$ & $10$\\
         VLCS \citep{fang2013unbiased} & $\{ \text{Caltech101}, \text{LabelMe}, \text{SUN09}, \text{VOC2007} \}$ & $10,729$ & $(3, 224, 224)$ & $5$\\
         PACS \citep{Li_2017_ICCV} & $\{ \text{art}, \text{cartoons}, \text{photos}, \text{sketches} \}$ & $9,991$ & $(3, 224, 224)$ & $7$\\
         OfficeHome \citep{venkateswara2017Deep} & $\{ \text{art}, \text{clipart}, \text{product}, \text{real} \}$ & $15,588$ & $(3, 224, 224)$ & $65$\\
         TerraIncognita \citep{beery2018recognition} & $\{ \text{L100}, \text{L38}, \text{L43}, \text{L46}\}$ & $24,788$ & $(3, 224, 224)$ & $10$\\
         DomainNet \citep{peng2019moment} & $\{ \text{clipart}, \text{infograph}, \text{painting}, \text{quickdraw}, \text{real}, \text{sketch} \}$ & $586,575$ & $(3, 224, 224)$ & $345$\\
         \bottomrule
    \end{tabular}}
    
    \label{tab:dg_datasets}
\end{table}

\paragraph{Compute Resources} We run the $10,560$ jobs on \textsc{NVIDIA GeForce RTX 2080 Ti} GPUs as well as \textsc{NVIDIA TITAN RTX} GPUs for the more resource intensive jobs.

\paragraph{Baseline models} We compare our algorithms to the following existing algorithms:

\begin{itemize}
    \item Empirical Risk Minimization (ERM, \cite{vapnik1998statistical}), where the sum of errors is minimized across domains.
    \item Group Distributionally Robust Optimization (DRO, \cite{sagawa2019distributionally}), where low performing domains are giving an increasing weight during training. 
    \item Inter-domain Mixup (Mixup, \cite{yan2020improve}).
    \item Meta-Learning for Domain Generalization (MLDG, \cite{li2018learning}). 
    \item Algorithms based on matching the latent distribution across domains:
    \begin{itemize}
        \item Domain-Adversarial Neural Networks (DANN, \cite{ganin2016domain}), where the distributional distance is an adversarial network.
        \item Class-conditional DANN (C-DANN, \cite{li2018deep}), which is a variant of DANN matching the class conditional distributions across domains.
        \item CORAL \cite{sun2016deep}, which aligns the mean and covariance of latent distributions. 
        \item MMD \cite{li2018domain}, which uses the MMD distance.
    \end{itemize}
    \item Invariant Risk Minimization (IRM \cite{arjovsky2019invariant}), which looks for a representation whose optimal linear classifier on top of the representation matches across domains.
    \item Style Agnostic Networks (SagNet, \cite{nam2021reducing}), which tries to reduce style bias of CNNs.
    \item Adaptive Risk Minimization (ARM, \cite{zhang2020adaptive}), which is based on meta-learning.
    \item Variance Risk Extrapolation (VREx, \cite{krueger2021out}), where they enforce the training risk to be similar across domains.
    \item Representation Self-Challenging (RSC, \cite{huang2020self}).
\end{itemize}

\lstset{
    language=Python,
    basicstyle=\ttfamily,
    keywordstyle=\color{blue}\ttfamily,
    stringstyle=\color{red}\ttfamily,
    commentstyle=\color{green}\ttfamily,
    morecomment=[l][\color{magenta}]{\#},
}

\paragraph{Implementation} To be more concrete, we change the code that computes the distributional distance penalty from this:

\newsavebox{\lsta}
\begin{lrbox}{\lsta}
\begin{lstlisting}
for i in range(nmb):
    for j in range(i + 1, nmb):
        penalty += self.dist_loss(features[i], features[j])

if nmb > 1:
    penalty /= (nmb * (nmb - 1) / 2)
\end{lstlisting}
\end{lrbox}

\scalebox{1.0}{\usebox{\lsta}}

to this: 

\newsavebox{\lstb}
\begin{lrbox}{\lstb}
\begin{lstlisting}
first = None
second = None

for i in range(nmb):
    slice = random.randint(0, len(features[i]))

    if first is None:
        first = features[i][:slice]
        second = features[i][slice:]
    else:
        first = torch.cat((first, features[i][:slice]), 0)
        second = torch.cat((second, features[i][slice:]), 0)

penalty = self.dist_loss(first, second)

\end{lstlisting}
\end{lrbox}

\scalebox{1.0}{\usebox{\lstb}}
.

Here is the concrete full class of our CausIRL with MMD model:

\begin{lstlisting}[language=Python]
class CausIRL_MMD(ERM):
    def __init__(self, input_shape, num_classes, num_domains, hparams):
        super(CausIRL_MMD, self).__init__(input_shape, num_classes, num_domains,
                                  hparams)
        self.kernel_type = "gaussian"

    def my_cdist(self, x1, x2):
        x1_norm = x1.pow(2).sum(dim=-1, keepdim=True)
        x2_norm = x2.pow(2).sum(dim=-1, keepdim=True)
        res = torch.addmm(x2_norm.transpose(-2, -1),
                          x1,
                          x2.transpose(-2, -1), alpha=-2).add_(x1_norm)
        return res.clamp_min_(1e-30)

    def gaussian_kernel(self, x, y, gamma=[0.001, 0.01, 0.1, 1, 10, 100,
                                           1000]):
        D = self.my_cdist(x, y)
        K = torch.zeros_like(D)

        for g in gamma:
            K.add_(torch.exp(D.mul(-g)))

        return K

    def mmd(self, x, y):
        Kxx = self.gaussian_kernel(x, x).mean()
        Kyy = self.gaussian_kernel(y, y).mean()
        Kxy = self.gaussian_kernel(x, y).mean()
        return Kxx + Kyy - 2 * Kxy
        

    def update(self, minibatches, unlabeled=None):
        objective = 0
        penalty = 0
        nmb = len(minibatches)

        features = [self.featurizer(xi) for xi, _ in minibatches]
        classifs = [self.classifier(fi) for fi in features]
        targets = [yi for _, yi in minibatches]

        first = None
        second = None

        for i in range(nmb):
            objective += F.cross_entropy(classifs[i] + 1e-16, targets[i])
            slice = random.randint(0, len(features[i]))
            if first is None:
                first = features[i][:slice]
                second = features[i][slice:]
            else:
                first = torch.cat((first, features[i][:slice]), 0)
                second = torch.cat((second, features[i][slice:]), 0)
        if len(first) > 1 and len(second) > 1:
            penalty = torch.nan_to_num(self.mmd(first, second))
        else:
            penalty = torch.tensor(0)
        
        objective /= nmb

        self.optimizer.zero_grad()
        (objective + (self.hparams['mmd_gamma']*penalty)).backward()
        self.optimizer.step()

        if torch.is_tensor(penalty):
            penalty = penalty.item()

        return {'loss': objective.item(), 'penalty': penalty}

\end{lstlisting}

\subsection{Model Selection: Training-Domain Validation Set}

We present here the results of our \DG experiments for the training-domain validation model selection method. Result are summarized in \cref{tab:dg_training}. For CausIRL with CORAL, the overall performance is slightly below vanilla CORAL. CausIRL with CORAL especially underperforms CORAL on the PACS dataset. On the other hand, CausIRL with CORAL performs better than CORAL on DomainNet. For CausIRL with MMD, the overall performance is significantly better than MMD. This overperformance is mainly driven by the results on TerraIncognita and DomainNet, where for the latter we observe a leap in accuracy from $23.4\%$ to $40.3\%$. %\stefan{is it really 40.3? I see 41.8?} 

\begin{table}[]
\caption{\DG experimental results for the training-domain validation selection method.}
    \begin{center}
\adjustbox{max width=\textwidth}{%
\begin{tabular}{lcccccccc}
\toprule
\textbf{Algorithm}        & \textbf{ColoredMNIST}     & \textbf{RotatedMNIST}     & \textbf{VLCS}             & \textbf{PACS}             & \textbf{OfficeHome}       & \textbf{TerraIncognita}   & \textbf{DomainNet}        & \textbf{Avg}              \\
\midrule
\rowcolor{lightblue}
\textbf{CausIRL with CORAL (ours)}                & 51.7 $\pm$ 0.1            & 97.9 $\pm$ 0.1            & 77.5 $\pm$ 0.6            & 85.8 $\pm$ 0.1            & 68.6 $\pm$ 0.3            & 47.3 $\pm$ 0.8            & 41.9 $\pm$ 0.1            & 67.3                      \\
CORAL                     & 51.5 $\pm$ 0.1            & 98.0 $\pm$ 0.1            & 78.8 $\pm$ 0.6            & 86.2 $\pm$ 0.3            & 68.7 $\pm$ 0.3            & 47.6 $\pm$ 1.0            & 41.5 $\pm$ 0.1            & 67.5                      \\
\rowcolor{lightblue}
\textbf{CausIRL with MMD (ours)}                 & 51.6 $\pm$ 0.1            & 97.9 $\pm$ 0.0            & 77.6 $\pm$ 0.4            & 84.0 $\pm$ 0.8            & 65.7 $\pm$ 0.6            & 46.3 $\pm$ 0.9            & 40.3 $\pm$ 0.2            & 66.2                      \\
MMD                       & 51.5 $\pm$ 0.2            & 97.9 $\pm$ 0.0            & 77.5 $\pm$ 0.9            & 84.6 $\pm$ 0.5            & 66.3 $\pm$ 0.1            & 42.2 $\pm$ 1.6            & 23.4 $\pm$ 9.5            & 63.3                      \\
\midrule
ERM                       & 51.5 $\pm$ 0.1            & 98.0 $\pm$ 0.0            & 77.5 $\pm$ 0.4            & 85.5 $\pm$ 0.2            & 66.5 $\pm$ 0.3            & 46.1 $\pm$ 1.8            & 40.9 $\pm$ 0.1            & 66.6                      \\
IRM                       & 52.0 $\pm$ 0.1            & 97.7 $\pm$ 0.1            & 78.5 $\pm$ 0.5            & 83.5 $\pm$ 0.8            & 64.3 $\pm$ 2.2            & 47.6 $\pm$ 0.8            & 33.9 $\pm$ 2.8            & 65.4                      \\
GroupDRO                  & 52.1 $\pm$ 0.0            & 98.0 $\pm$ 0.0            & 76.7 $\pm$ 0.6            & 84.4 $\pm$ 0.8            & 66.0 $\pm$ 0.7            & 43.2 $\pm$ 1.1            & 33.3 $\pm$ 0.2            & 64.8                      \\
Mixup                     & 52.1 $\pm$ 0.2            & 98.0 $\pm$ 0.1            & 77.4 $\pm$ 0.6            & 84.6 $\pm$ 0.6            & 68.1 $\pm$ 0.3            & 47.9 $\pm$ 0.8            & 39.2 $\pm$ 0.1            & 66.7                      \\
MLDG                      & 51.5 $\pm$ 0.1            & 97.9 $\pm$ 0.0            & 77.2 $\pm$ 0.4            & 84.9 $\pm$ 1.0            & 66.8 $\pm$ 0.6            & 47.7 $\pm$ 0.9            & 41.2 $\pm$ 0.1            & 66.7                      \\
DANN                      & 51.5 $\pm$ 0.3            & 97.8 $\pm$ 0.1            & 78.6 $\pm$ 0.4            & 83.6 $\pm$ 0.4            & 65.9 $\pm$ 0.6            & 46.7 $\pm$ 0.5            & 38.3 $\pm$ 0.1            & 66.1                      \\
CDANN                     & 51.7 $\pm$ 0.1            & 97.9 $\pm$ 0.1            & 77.5 $\pm$ 0.1            & 82.6 $\pm$ 0.9            & 65.8 $\pm$ 1.3            & 45.8 $\pm$ 1.6            & 38.3 $\pm$ 0.3            & 65.6                      \\
MTL                       & 51.4 $\pm$ 0.1            & 97.9 $\pm$ 0.0            & 77.2 $\pm$ 0.4            & 84.6 $\pm$ 0.5            & 66.4 $\pm$ 0.5            & 45.6 $\pm$ 1.2            & 40.6 $\pm$ 0.1            & 66.2                      \\
SagNet                    & 51.7 $\pm$ 0.0            & 98.0 $\pm$ 0.0            & 77.8 $\pm$ 0.5            & 86.3 $\pm$ 0.2            & 68.1 $\pm$ 0.1            & 48.6 $\pm$ 1.0            & 40.3 $\pm$ 0.1            & 67.2                      \\
ARM                       & 56.2 $\pm$ 0.2            & 98.2 $\pm$ 0.1            & 77.6 $\pm$ 0.3            & 85.1 $\pm$ 0.4            & 64.8 $\pm$ 0.3            & 45.5 $\pm$ 0.3            & 35.5 $\pm$ 0.2            & 66.1                      \\
VREx                      & 51.8 $\pm$ 0.1            & 97.9 $\pm$ 0.1            & 78.3 $\pm$ 0.2            & 84.9 $\pm$ 0.6            & 66.4 $\pm$ 0.6            & 46.4 $\pm$ 0.6            & 33.6 $\pm$ 2.9            & 65.6                      \\
RSC                       & 51.7 $\pm$ 0.2            & 97.6 $\pm$ 0.1            & 77.1 $\pm$ 0.5            & 85.2 $\pm$ 0.9            & 65.5 $\pm$ 0.9            & 46.6 $\pm$ 1.0            & 38.9 $\pm$ 0.5            & 66.1                      \\
\bottomrule
\end{tabular}}
\end{center}
    
    \label{tab:dg_training}
\end{table}

\subsection{Model selection: leave-one-domain-out cross-validation}

We here present the complete results for the leave-one-domain-out cross-validation model selection method in \cref{tab:dg_leave2}.

\begin{table}[]
\caption{\DG experimental results for the leave-one-domain-out cross-validation model selection method.}
    \begin{center}
\adjustbox{max width=\textwidth}{%
\begin{tabular}{lcccccccc}
\toprule
\textbf{Algorithm}        & \textbf{ColoredMNIST}     & \textbf{RotatedMNIST}     & \textbf{VLCS}             & \textbf{PACS}             & \textbf{OfficeHome}       & \textbf{TerraIncognita}   & \textbf{DomainNet}        & \textbf{Avg}              \\
\midrule
\rowcolor{lightblue}
\textbf{CausIRL with CORAL (ours)}                & 39.1 $\pm$ 2.0            & 97.8 $\pm$ 0.1            & 76.5 $\pm$ 1.0            & 83.6 $\pm$ 1.2            & 68.1 $\pm$ 0.3            & 47.4 $\pm$ 0.5            & 41.8 $\pm$ 0.1            & 64.9                      \\
CORAL                     & 39.7 $\pm$ 2.8            & 97.8 $\pm$ 0.1            & 78.7 $\pm$ 0.4            & 82.6 $\pm$ 0.5            & 68.5 $\pm$ 0.2            & 46.3 $\pm$ 1.7            & 41.1 $\pm$ 0.1            & 65.0                      \\
\rowcolor{lightblue}
\textbf{CausIRL with MMD (ours)}                  & 36.9 $\pm$ 0.2            & 97.6 $\pm$ 0.1            & 78.2 $\pm$ 0.9            & 84.0 $\pm$ 0.9            & 65.1 $\pm$ 0.7            & 47.9 $\pm$ 0.3            & 38.9 $\pm$ 0.8            & 64.1                      \\
MMD                       & 36.8 $\pm$ 0.1            & 97.8 $\pm$ 0.1            & 77.3 $\pm$ 0.5            & 83.2 $\pm$ 0.2            & 60.2 $\pm$ 5.2            & 46.5 $\pm$ 1.5            & 23.4 $\pm$ 9.5            & 60.7                      \\
\midrule
ERM                       & 36.7 $\pm$ 0.1            & 97.7 $\pm$ 0.0            & 77.2 $\pm$ 0.4            & 83.0 $\pm$ 0.7            & 65.7 $\pm$ 0.5            & 41.4 $\pm$ 1.4            & 40.6 $\pm$ 0.2            & 63.2                      \\
IRM                       & 40.3 $\pm$ 4.2            & 97.0 $\pm$ 0.2            & 76.3 $\pm$ 0.6            & 81.5 $\pm$ 0.8            & 64.3 $\pm$ 1.5            & 41.2 $\pm$ 3.6            & 33.5 $\pm$ 3.0            & 62.0                      \\
GroupDRO                  & 36.8 $\pm$ 0.1            & 97.6 $\pm$ 0.1            & 77.9 $\pm$ 0.5            & 83.5 $\pm$ 0.2            & 65.2 $\pm$ 0.2            & 44.9 $\pm$ 1.4            & 33.0 $\pm$ 0.3            & 62.7                      \\
Mixup                     & 33.4 $\pm$ 4.7            & 97.8 $\pm$ 0.0            & 77.7 $\pm$ 0.6            & 83.2 $\pm$ 0.4            & 67.0 $\pm$ 0.2            & 48.7 $\pm$ 0.4            & 38.5 $\pm$ 0.3            & 63.8                      \\
MLDG                      & 36.7 $\pm$ 0.2            & 97.6 $\pm$ 0.0            & 77.2 $\pm$ 0.9            & 82.9 $\pm$ 1.7            & 66.1 $\pm$ 0.5            & 46.2 $\pm$ 0.9            & 41.0 $\pm$ 0.2            & 64.0                      \\
DANN                      & 40.7 $\pm$ 2.3            & 97.6 $\pm$ 0.2            & 76.9 $\pm$ 0.4            & 81.0 $\pm$ 1.1            & 64.9 $\pm$ 1.2            & 44.4 $\pm$ 1.1            & 38.2 $\pm$ 0.2            & 63.4                      \\
CDANN                     & 39.1 $\pm$ 4.4            & 97.5 $\pm$ 0.2            & 77.5 $\pm$ 0.2            & 78.8 $\pm$ 2.2            & 64.3 $\pm$ 1.7            & 39.9 $\pm$ 3.2            & 38.0 $\pm$ 0.1            & 62.2                      \\
MTL                       & 35.0 $\pm$ 1.7            & 97.8 $\pm$ 0.1            & 76.6 $\pm$ 0.5            & 83.7 $\pm$ 0.4            & 65.7 $\pm$ 0.5            & 44.9 $\pm$ 1.2            & 40.6 $\pm$ 0.1            & 63.5                      \\
SagNet                    & 36.5 $\pm$ 0.1            & 94.0 $\pm$ 3.0            & 77.5 $\pm$ 0.3            & 82.3 $\pm$ 0.1            & 67.6 $\pm$ 0.3            & 47.2 $\pm$ 0.9            & 40.2 $\pm$ 0.2            & 63.6                      \\
ARM                       & 36.8 $\pm$ 0.0            & 98.1 $\pm$ 0.1            & 76.6 $\pm$ 0.5            & 81.7 $\pm$ 0.2            & 64.4 $\pm$ 0.2            & 42.6 $\pm$ 2.7            & 35.2 $\pm$ 0.1            & 62.2                      \\
VREx                      & 36.9 $\pm$ 0.3            & 93.6 $\pm$ 3.4            & 76.7 $\pm$ 1.0            & 81.3 $\pm$ 0.9            & 64.9 $\pm$ 1.3            & 37.3 $\pm$ 3.0            & 33.4 $\pm$ 3.1            & 60.6                      \\
RSC                       & 36.5 $\pm$ 0.2            & 97.6 $\pm$ 0.1            & 77.5 $\pm$ 0.5            & 82.6 $\pm$ 0.7            & 65.8 $\pm$ 0.7            & 40.0 $\pm$ 0.8            & 38.9 $\pm$ 0.5            & 62.7                      \\
\bottomrule
\end{tabular}}
\end{center}
    
    \label{tab:dg_leave2}
\end{table}

\subsection{Model Selection: Test-Domain Validation Set (Oracle)}

We here present the complete results for the test-domain validation set model selection method in \cref{tab:dg_oracle2}.

%\stefan{Between table 2,3,4 only keep one experiment, describe the rest and then put the tables in the appendix and reference it}

\begin{table}[]
\caption{\DG experimental results for the test-domain validation set model selection method.}
    \begin{center}
\adjustbox{max width=\textwidth}{%
\begin{tabular}{lcccccccc}
\toprule
\textbf{Algorithm}        & \textbf{ColoredMNIST}     & \textbf{RotatedMNIST}     & \textbf{VLCS}             & \textbf{PACS}             & \textbf{OfficeHome}       & \textbf{TerraIncognita}   & \textbf{DomainNet}        & \textbf{Avg}              \\
\midrule
\rowcolor{lightblue}
\textbf{CausIRL with CORAL (ours)}                & 58.4 $\pm$ 0.3            & 98.0 $\pm$ 0.1            & 78.2 $\pm$ 0.1            & 87.6 $\pm$ 0.1            & 67.7 $\pm$ 0.2            & 53.4 $\pm$ 0.4            & 42.1 $\pm$ 0.1            & 69.4                      \\
CORAL                     & 58.6 $\pm$ 0.5            & 98.0 $\pm$ 0.0            & 77.7 $\pm$ 0.2            & 87.1 $\pm$ 0.5            & 68.4 $\pm$ 0.2            & 52.8 $\pm$ 0.2            & 41.8 $\pm$ 0.1            & 69.2                      \\
\rowcolor{lightblue}
\textbf{CausIRL with MMD (ours)}                  & 63.7 $\pm$ 0.8            & 97.9 $\pm$ 0.1            & 78.1 $\pm$ 0.1            & 86.6 $\pm$ 0.7            & 65.2 $\pm$ 0.6            & 52.2 $\pm$ 0.3            & 40.6 $\pm$ 0.2            & 69.2                      \\
MMD                       & 63.3 $\pm$ 1.3            & 98.0 $\pm$ 0.1            & 77.9 $\pm$ 0.1            & 87.2 $\pm$ 0.1            & 66.2 $\pm$ 0.3            & 52.0 $\pm$ 0.4            & 23.5 $\pm$ 9.4            & 66.9                      \\
\midrule
ERM                       & 57.8 $\pm$ 0.2            & 97.8 $\pm$ 0.1            & 77.6 $\pm$ 0.3            & 86.7 $\pm$ 0.3            & 66.4 $\pm$ 0.5            & 53.0 $\pm$ 0.3            & 41.3 $\pm$ 0.1            & 68.7                      \\
IRM                       & 67.7 $\pm$ 1.2            & 97.5 $\pm$ 0.2            & 76.9 $\pm$ 0.6            & 84.5 $\pm$ 1.1            & 63.0 $\pm$ 2.7            & 50.5 $\pm$ 0.7            & 28.0 $\pm$ 5.1            & 66.9                      \\
GroupDRO                  & 61.1 $\pm$ 0.9            & 97.9 $\pm$ 0.1            & 77.4 $\pm$ 0.5            & 87.1 $\pm$ 0.1            & 66.2 $\pm$ 0.6            & 52.4 $\pm$ 0.1            & 33.4 $\pm$ 0.3            & 67.9                      \\
Mixup                     & 58.4 $\pm$ 0.2            & 98.0 $\pm$ 0.1            & 78.1 $\pm$ 0.3            & 86.8 $\pm$ 0.3            & 68.0 $\pm$ 0.2            & 54.4 $\pm$ 0.3            & 39.6 $\pm$ 0.1            & 69.0                      \\
MLDG                      & 58.2 $\pm$ 0.4            & 97.8 $\pm$ 0.1            & 77.5 $\pm$ 0.1            & 86.8 $\pm$ 0.4            & 66.6 $\pm$ 0.3            & 52.0 $\pm$ 0.1            & 41.6 $\pm$ 0.1            & 68.7                      \\
DANN                      & 57.0 $\pm$ 1.0            & 97.9 $\pm$ 0.1            & 79.7 $\pm$ 0.5            & 85.2 $\pm$ 0.2            & 65.3 $\pm$ 0.8            & 50.6 $\pm$ 0.4            & 38.3 $\pm$ 0.1            & 67.7                      \\
CDANN                     & 59.5 $\pm$ 2.0            & 97.9 $\pm$ 0.0            & 79.9 $\pm$ 0.2            & 85.8 $\pm$ 0.8            & 65.3 $\pm$ 0.5            & 50.8 $\pm$ 0.6            & 38.5 $\pm$ 0.2            & 68.2                      \\
MTL                       & 57.6 $\pm$ 0.3            & 97.9 $\pm$ 0.1            & 77.7 $\pm$ 0.5            & 86.7 $\pm$ 0.2            & 66.5 $\pm$ 0.4            & 52.2 $\pm$ 0.4            & 40.8 $\pm$ 0.1            & 68.5                      \\
SagNet                    & 58.2 $\pm$ 0.3            & 97.9 $\pm$ 0.0            & 77.6 $\pm$ 0.1            & 86.4 $\pm$ 0.4            & 67.5 $\pm$ 0.2            & 52.5 $\pm$ 0.4            & 40.8 $\pm$ 0.2            & 68.7                      \\
ARM                       & 63.2 $\pm$ 0.7            & 98.1 $\pm$ 0.1            & 77.8 $\pm$ 0.3            & 85.8 $\pm$ 0.2            & 64.8 $\pm$ 0.4            & 51.2 $\pm$ 0.5            & 36.0 $\pm$ 0.2            & 68.1                      \\
VREx                      & 67.0 $\pm$ 1.3            & 97.9 $\pm$ 0.1            & 78.1 $\pm$ 0.2            & 87.2 $\pm$ 0.6            & 65.7 $\pm$ 0.3            & 51.4 $\pm$ 0.5            & 30.1 $\pm$ 3.7            & 68.2                      \\
RSC                       & 58.5 $\pm$ 0.5            & 97.6 $\pm$ 0.1            & 77.8 $\pm$ 0.6            & 86.2 $\pm$ 0.5            & 66.5 $\pm$ 0.6            & 52.1 $\pm$ 0.2            & 38.9 $\pm$ 0.6            & 68.2                      \\
\bottomrule
\end{tabular}}
\end{center}
    
    \label{tab:dg_oracle2}
\end{table}

\end{document}